\documentclass{article}
\usepackage[final]{neurips_2021}
\usepackage[utf8]{inputenc} 
\usepackage[T1]{fontenc}    
\usepackage[hidelinks]{hyperref}       
\usepackage{url}            
\usepackage{booktabs}       
\usepackage[table]{xcolor}
\newcolumntype{K}{!{\color{white}\ }c}

\usepackage{amsfonts}       
\usepackage{nicefrac}       
\usepackage{microtype}      
\usepackage{xcolor}         
\usepackage{graphicx}
\usepackage{enumitem}
\usepackage{amsmath}
\usepackage{pifont}
\usepackage{placeins}
\usepackage{braket}
\usepackage{soul}
\usepackage{enumitem}
\usepackage{subcaption}
\usepackage{algorithm, algorithmic}

\usepackage[colorinlistoftodos,prependcaption,textsize=tiny]{todonotes}
\usepackage{bm, bbm}

\author{
Ryan Cory-Wright\thanks{Corresponding Author}\\
Department of Analytics, Marketing and Operations, Imperial College Business School, London, UK\\
\texttt{r.cory-wright@imperial.ac.uk}\\
 \And Cristina Cornelio\\
 Samsung AI, Cambridge, UK\\
 \texttt{c.cornelio@samsung.com}
\And
Sanjeeb Dash\\
IBM Thomas J. Watson Research Center\\
Yorktown Heights, USA\\
\texttt{sanjeebd@us.ibm.com}
\And
Bachir El Khadir\\
IBM Thomas J. Watson Research Center\\
Yorktown Heights, USA\\
\texttt{bachir009@gmail.com}
\And 
Lior Horesh\\
IBM Thomas J. Watson Research Center\\
Yorktown Heights, USA\\
\texttt{lhoresh@us.ibm.com}
}

\newtheorem{theorem}{Theorem}
\title{\color{black} Evolving Scientific Discovery by Unifying Data and Background Knowledge with AI Hilbert} 

\newtheorem{example}{Example}

\begin{document}

\newcommand{\xmark}{\ding{55}}%
\maketitle

\begin{abstract}
The discovery of scientific formulae that parsimoniously explain natural phenomena and align with existing background theory is a key goal in science. Historically, scientists have derived natural laws by manipulating equations based on existing knowledge, forming new equations, and verifying them experimentally. In recent years, data-driven scientific discovery has emerged as a viable competitor in settings with large amounts of experimental data. Unfortunately, data-driven methods often fail to discover valid laws when data is noisy or scarce. Accordingly, recent works combine regression and reasoning to eliminate formulae inconsistent with background theory. However, the problem of searching over the space of formulae consistent with background theory to find one that {\color{black}best fits the data}
is not well-solved. We propose a solution to this problem when all axioms and scientific laws are expressible via polynomial equalities and inequalities and argue that our approach is widely applicable. We 
model notions of minimal complexity using binary variables and logical constraints, solve polynomial optimization problems via mixed-integer linear or semidefinite optimization, and prove the validity of our scientific discoveries in a principled manner using Positivstellensatz certificates. 
The optimization techniques leveraged in this paper allow our approach to run in polynomial time with fully correct background theory {\color{black}(under an assumption that the complexity of our derivation is bounded)}, or non-deterministic polynomial (NP) time with partially correct background theory. We demonstrate that some famous scientific laws, including Kepler’s Third Law of Planetary Motion, the Hagen-Poiseuille Equation, and the Radiated Gravitational Wave Power equation, can be derived in a principled manner from 
axioms and experimental data.
\end{abstract}
\newpage

\section{Introduction}\label{sec:intro}
\begin{figure}[b!]
    \centering
    \includegraphics[scale=0.6]{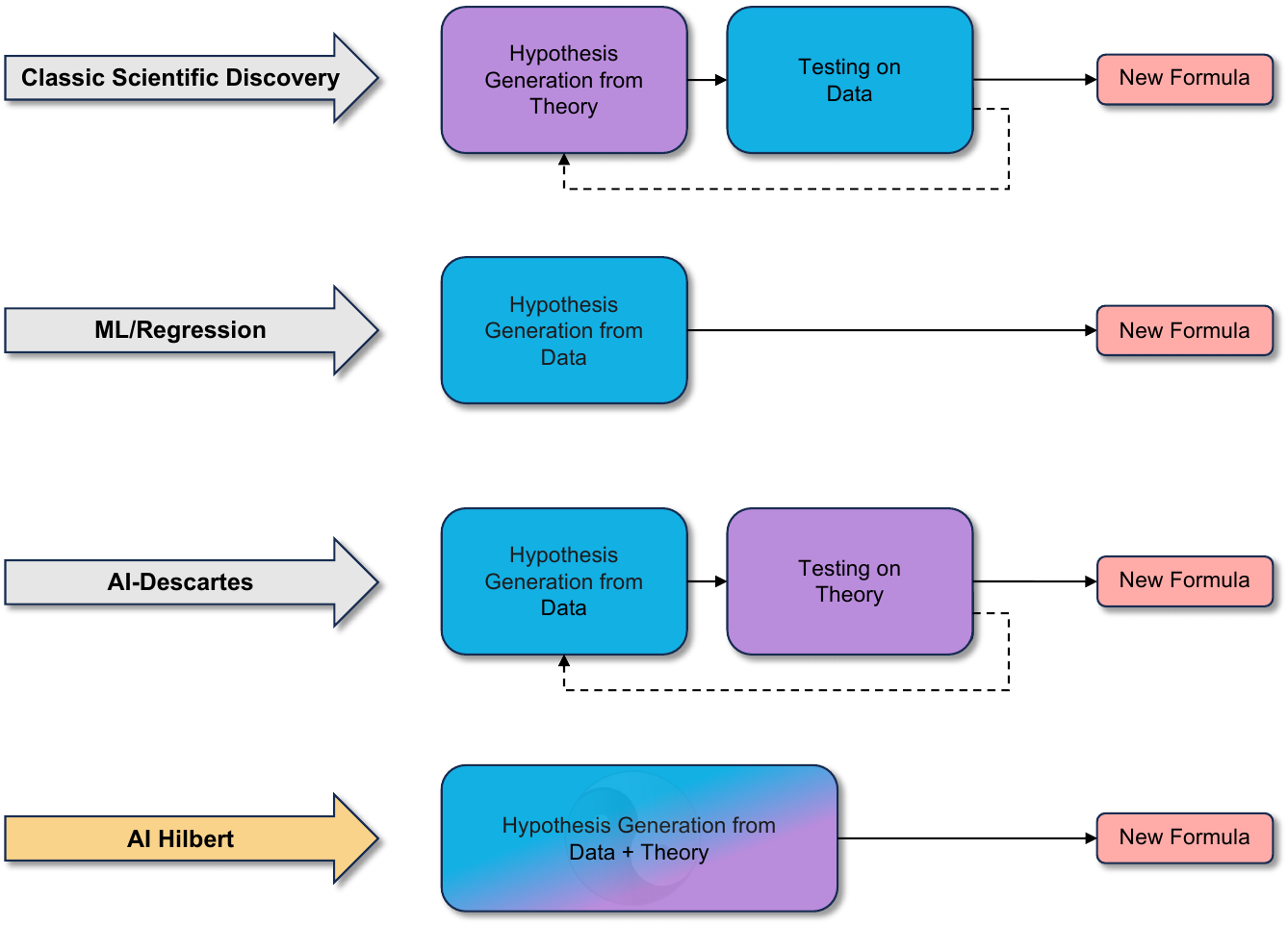}
    \caption{\textbf{
    Comparison of scientific discovery paradigms.} %
     Traditional scientific discovery formulates hypotheses using existing theory and observed phenomena. These hypotheses are 
     validated and tested using data. 
     In contrast, machine learning 
     methods rely on large datasets to identify patterns. 
     AI-Descartes~\cite{cornelio2021ai} proposes an inversion of the conventional scientific discovery paradigm. It 
     generates 
     hypotheses from observed data 
     and 
     validates 
     them against 
     known theory. 
     However, in AI-Descartes, theory and data remain disjoint and do not mutually enhance one another.
     In contrast, our work, AI Hilbert, combines data and theory to formulate hypotheses. Unlike conventional methods, insights in data and knowledge embedded in the theory collaboratively reduce the search space. These two components complement each other: theory compensates for noisy or sparse data, while data compensates for inconsistent or incomplete theory. Note that blue denotes components associated with data, purple denotes components linked to theory, and dashed lines represent iterative processes.}
    \label{fig:alternatives_scientific_method}
\end{figure}

A fundamental problem in science 
involves explaining natural phenomena in a manner consistent with noisy experimental data and a body of potentially inexact and incomplete background knowledge about the universe's laws \cite{de2020understanding}. In the past few centuries, The Scientific Method \citep{simon1973does} has led to significant progress in discovering new laws. Unfortunately, the rate of emergence of these laws and their contribution to economic growth is stagnating relative to the amount of capital invested in deducing them \cite{brynjolfsson2018artificial,bhattacharya2020stagnation}. Indeed, Dirac~\cite{dirac1978directions} noted that it is now more challenging for first-rate physicists to make second-rate discoveries than it was previously for second-rate physicists to make first-rate ones, while Arora et al.~\cite{arora2018decline} found that the marginal value of scientific discoveries to large companies has declined since the fall of the Berlin Wall{\color{black}. Moreover, Bloom et al.~\cite{bloom2020ideas} have found that research productivity in the United States halves every thirteen years because good scientific ideas are getting harder to find}.
This phenomenon can be partly explained by analogy to 
\cite{cowen2011great}{\color{black}:} 
The Scientific Method has picked most of the ``low-hanging fruit'' in science{\color{black},} 
such as natural laws that relate physical quantities using a small number of low-degree polynomials. This calls for more disciplined and principled alternatives to The Scientific Method, which integrate background information and experimental data to generate and verify higher dimensional laws of nature, thereby promoting scientific discovery \citep[c.f.][]{kitano2021nobel, wang2023scientific}.
{\color{black}Accordingly, Figure~\ref{fig:alternatives_scientific_method} provides an overview of these alternatives.}

{\color{black}Even as the rate of scientific discovery has decreased, the scalability of global optimization methods has significantly improved. Indeed, as we argue in this paper, global optimization methods are now a mature technology capable of searching over the space of scientific laws — owing to Moore's law and significant theoretical and computational advances by the optimization community \citep[see][for reviews]{bixby2007progress, gupta2022branch, bertsimas2023matrix}. }{\color{black}Indeed, }Bertsimas and Dunn~\citep[][Chap. 1]{bertsimas2019machine} observed that the speedup in raw computing power between $1991$ and $2015$ is at least six orders of magnitude. Additionally, 
polynomial optimization has become much more scalable since the works of {\color{black} Lasserre~\cite{lasserre2001global} and }Parrilo~\cite{parrilo2003semidefinite}, and primal-dual interior-point methods~\cite{nesterov1994interior, renegar2001mathematical, skajaa2015homogeneous} have improved considerably, with excellent implementations now available in, for example, the \verb|Mosek| solver~\cite{andersen2000mosek}. {\color{black}Indeed, even methods for non-convex quadratically constrained problems have achieved machine-independent speedups of nearly $200$ since their integration within commercial solvers in $2019$ \cite{achterberg2019s, gupta2022branch}.}

In this paper, we propose a new approach to scientific discovery that leverages these advances by the optimization community. Given a set of background axioms, theorems, and laws expressible as a basic semialgebraic set (i.e., a system of polynomial equalities and inequalities) and observations from experimental data, we derive new laws representable as polynomial expressions that are either exactly or approximately consistent with existing laws and experimental data by solving polynomial optimization problems via linear and semidefinite optimization. By leveraging fundamental results from real algebraic geometry, we obtain formal proofs of the correctness of our laws as a byproduct of the optimization problems. This is notable, because existing automated approaches to scientific discovery, as reviewed in Section \ref{sec:litrev}, often rely upon deep learning techniques that do not provide formal proofs and are prone to ``hallucinating'' incorrect scientific laws that cannot be automatically proven or disproven, analogously to output from state-of-the-art Large Language Models such as GPT-$4$ \cite{open2023gpt4}. As such, any new laws derived by these systems cannot easily be explained or justified. 

{\color{black}Conversely}, our approach discovers new scientific laws by solving an optimization problem to minimize a weighted sum of discrepancies between the proposed law and experimental data, plus the distance between the {\color{black} proposed} law and its projection onto the set of symbolic laws derivable from background theory. As a result, our approach discovers scientific laws alongside a proof of their consistency with existing background theory \textit{by default}. Moreover, our approach is scalable; it runs in polynomial time {\color{black}with respect to the number of symbolic variables and axioms} (when the degree of the polynomial certificates we search over is bounded; see Section \ref{ssec:overallproblem}) with a complete and consistent background theory.

We believe our approach could be a first step towards discovering new laws of the universe which involve higher degree polynomials and are impractical for scientists to discover without the aid of modern {\color{black}optimization} solvers and high-performance computing environments. Further, our approach is potentially useful for reconciling mutually inconsistent axioms. Indeed, if a system of scientific laws is mutually inconsistent (in the sense that no point satisfies all laws simultaneously), our polynomial optimization problem offers a formal proof of its inconsistency. 

\subsection{Literature Review}\label{sec:litrev}
We propose an approach to scientific discovery, which we term \verb|AI-Hilbert|, that uses polynomial optimization to obtain scientific formulae derivable from background theory axioms and consistent with experimental data. 
This differs from existing works on scientific discovery that express prior knowledge as constraints on the functional form of a learned model (e.g., shape constraints such as monotonicity \cite{curmei2020shape}). Indeed, shape-constrained approaches to scientific discovery have been proposed \citep[][]{kubalik2020symbolic, kubalik2021multi,engle2022deterministic,  curmei2020shape, bertsimas2023learning}, while discovering scientific laws that are simultaneously derivable from prior knowledge expressed as polynomials and experimental data is{\color{black}, to our knowledge, a new approach.}

\verb|AI-Hilbert| builds upon two areas {\color{black}of the optimization and discovery literature} typically considered in isolation: sum-of-squares optimization techniques for solving polynomial optimization problems, and data-driven techniques for symbolic discovery. We now review the relevant literature.

\paragraph{Sum-of-Squares Optimization:}
Sum-of-squares optimization has been an important component of global optimization methods since the seminal work of Parrilo~\cite{parrilo2003semidefinite} (see also Lasserre~\cite{lasserre2001global}), which combines two key observations. First, sum-of-squares decompositions of multivariate polynomials can be computed via semidefinite optimization, so optimizing over sum-of-squares polynomials is no harder than performing semidefinite optimization. Second, owing to a fundamental result from real algebraic geometry, 
the Positivstellensatz~\cite{krivine1964anneaux, stengle1974nullstellensatz, putinar1993positive}, polynomials of bounded degree defined on basic semialgebraic sets can be certified as non-negative over these sets by representing them as systems of sum-of-squares polynomials (see Section \ref{ssec:notation}). Consequently, optimizing over a real polynomial system is 
equivalent to solving a (larger) sum-of-squares optimization problem, and thus a tractable convex problem. These observations have allowed an entire field of optimization to blossom; see Blekherman et al.~\cite{blekherman2012semidefinite}, Hall~\cite{hall2019engineering} for reviews. However, to our knowledge, no works have proposed using sum-of-squares optimization to discover scientific formulae. The closest works are Clegg et al.~\cite{clegg1996using}, who propose using Gr{\"o}bner bases to design proofs of unsatisfiability, Curmei and Hall~\cite{curmei2020shape}, who propose a sum-of-squares approach to fitting a polynomial to data under very general constraints on the functional form of the polynomial, e.g., non-negativity of the derivative over a box, Ahmadi and El Khadir~\cite{bachir2023sideinfo}, who propose learning the behavior of noisy dynamical systems via semialgebraic techniques, and Fawzi et al.~\cite{fawzi2019learning}, who propose learning proofs of optimality of stable set problems by combining reinforcement learning with the Positivstellensatz. However, determining whether polynomial optimization is 
useful for scientific discovery {\color{black}is, to our knowledge,} open.

\paragraph{Data-Driven Approaches to Scientific Discovery:}
The availability of large amounts of scientific data generated and collected over the past few decades has spurred increasing interest in data-driven methods for scientific discovery that aim to identify symbolic equations that accurately explain high-dimensional datasets. Bongard and Lipson~\cite{bongard2007automated} and Schmidt and Lipson~\cite{schmidt2009distilling} proposed using heuristics and genetic programming to discover scientifically meaningful formulae, and implement their approach in the \verb|Eureqa| software system~\cite{dubvcakova2011eureqa}. Other proposed approaches are based on mixed-integer global optimization~\cite{austel2017globally, cozad2018global}, sparse regression~\cite{brunton2016discovering, rudy2017data, bertsimas2023learning}, Cylindrical Algebraic Decomposition~\cite{fulton2015keymaera}, neural networks~\cite{iten2020discovering, landajuela2022unified}, and Bayesian Markov Chain Monte Carlo approaches~\cite{guimera2020bayesian}. See \cite{karagiorgi2022machine, baum2021artificial} for reviews of data-driven scientific discovery in fundamental physics and chemistry.

Data-driven approaches have 
been shown by several authors to perform well in highly overdetermined settings with limited amounts of noise. For instance, Udrescu et al.~\cite{udrescu2020ai, udrescu2020ai2} proposed a method called \verb|AI-Feynman|, which combines neural networks with physics-based techniques to discover symbolic formulae. Moreover, they constructed a benchmark dataset of $100$ scientific laws derived from Richard Feynman's lecture notes~\cite{feynman1965feynman}, with $100,000$ noiseless experimental observations of each scientific law, and demonstrated that while the \verb|Eurequa| system could recover an already impressive $71/100$ instances from the data, their approach could recover all one hundred; see 
Cornelio et al.~\cite{cornelio2021ai} for a review of scientific discovery systems.

Unfortunately, data-driven approaches to scientific discovery have at least three significant drawbacks. First, they are not data efficient \cite{fujinuma2022big} and only reliably recover scientific formulae in 
overdetermined settings with 
orders of magnitude more data than a human would likely need to make the same discoveries. Indeed, Matsubara et al.~\cite{matsubara2022srsd} recently argued that the sampling regime used by \verb|AI-Feynman| is unrealistic, because it samples values far from those observable in the real world. Moreover, Cornelio et al.~\cite{cornelio2021ai} recently benchmarked \verb|AI-Feyman| on $81$ of the $100$ 
laws, but with $10$ (rather than $100,000$) observations per law, and where each experimental observation is contaminated with a small amount of noise. In this limited data setting, Cornelio et al.~\cite{cornelio2021ai} found that \verb|AI-Feynman| recovered $40$ of the $81$ laws considered, whereas {\color{black}\cite{cornelio2021ai}} 
were able to recover $49/81$ laws using their symbolic regression solver.  This performance degradation is a significant issue in practice because scientific data is typically expensive to obtain and scarce and noisy. Second, purely data-driven methods are agnostic to important background information, such as existing literature, that valid scientific formulae should be consistent with unless there is extraordinary experimental evidence that the literature is incorrect. This implies that data-driven methods search over a larger space of laws than is necessary, require more data than a human would need to derive a valid law, and frequently propose laws that are not scientifically meaningful. Third, {\color{black}data-driven methods} typically do not provide interpretable explanations for why their discoveries are valid \citep[c.f.][]{rudin2019stop}, which makes diagnosing whether their discoveries are consistent with existing theory challenging.

To account for background theory in scientific discovery, Cornelio et al.~\cite{cornelio2021ai} recently proposed an approach called \verb|AI-Descartes|, which iteratively generates plausible scientific formulae using a mixed-integer nonlinear symbolic regression solver \citep[see also][]{austel2017globally}, and tests whether these formulae are derivable from the background knowledge. {\color{black}If} they are not, the method provides a set of reasoning-based measures to compute how distant the formulae {induced} from the data are from the background theory but is unable to recover the correct formulae.
 This is because their approach {induces} potential scientific laws from data and subsequently tests the hypothesis against the background theory, rather than learning from axioms and data simultaneously.

\subsection{Contributions and Structure}

{\color{black}
We propose a new paradigm for scientific discovery that derives polynomial laws simultaneously consistent with experimental data and a body of background knowledge expressible as polynomial equalities and inequalities. We term our approach \verb|AI-Hilbert|, inspired by 
the work of David Hilbert, one of the first mathematicians to investigate the relationship between sum-of-squares and non-negative polynomials \cite{hilbert2019mathematical}. 
}

Our approach automatically provides an axiomatic derivation of the correctness of a discovered scientific law, conditional on the correctness of our background theory. Moreover, in instances with inconsistent background theory, our approach can successfully identify the sources of inconsistency by performing best subset selection to determine the axioms that best explain the data. This is notably different from current data-driven approaches to scientific discovery, which often generate spurious laws in limited data settings and fail to differentiate between valid and invalid discoveries or provide explanations of their derivations. We illustrate our approach by axiomatically deriving some of the most frequently cited natural laws in the scientific literature, including Kepler's Third Law and Einstein's Relativistic Time Dilation Law, among other scientific discoveries.

A second contribution of our approach is that it permits fine-grained control of the tractability of the scientific discovery process, by bounding the degree of the coefficients in the Positivstellensatz certificates that are searched over (see Section \ref{ssec:notation}, for a formal statement of the Positivstellensatz). This differs from prior works on automated scientific discovery, which offers more limited control over their time complexity. For instance, in the special case of scientific discovery with a complete body of background theory and no experimental data, to our knowledge, the only current alternative to our approach is symbolic regression \citep[see, e.g.,][]{cozad2018global}, which requires genetic programming or mixed-integer nonlinear programming techniques that are not guaranteed to run in polynomial time. On the other hand, our approach searches for polynomial certificates of a bounded degree {\color{black}via} 
a fixed level of the sum-of-squares hierarchy \cite{lasserre2001global, parrilo2003semidefinite}, which can be searched over in polynomial time \cite{nesterov1994interior, ramana1997exact}.

\begin{figure}
    \centering
\includegraphics[scale=0.52]{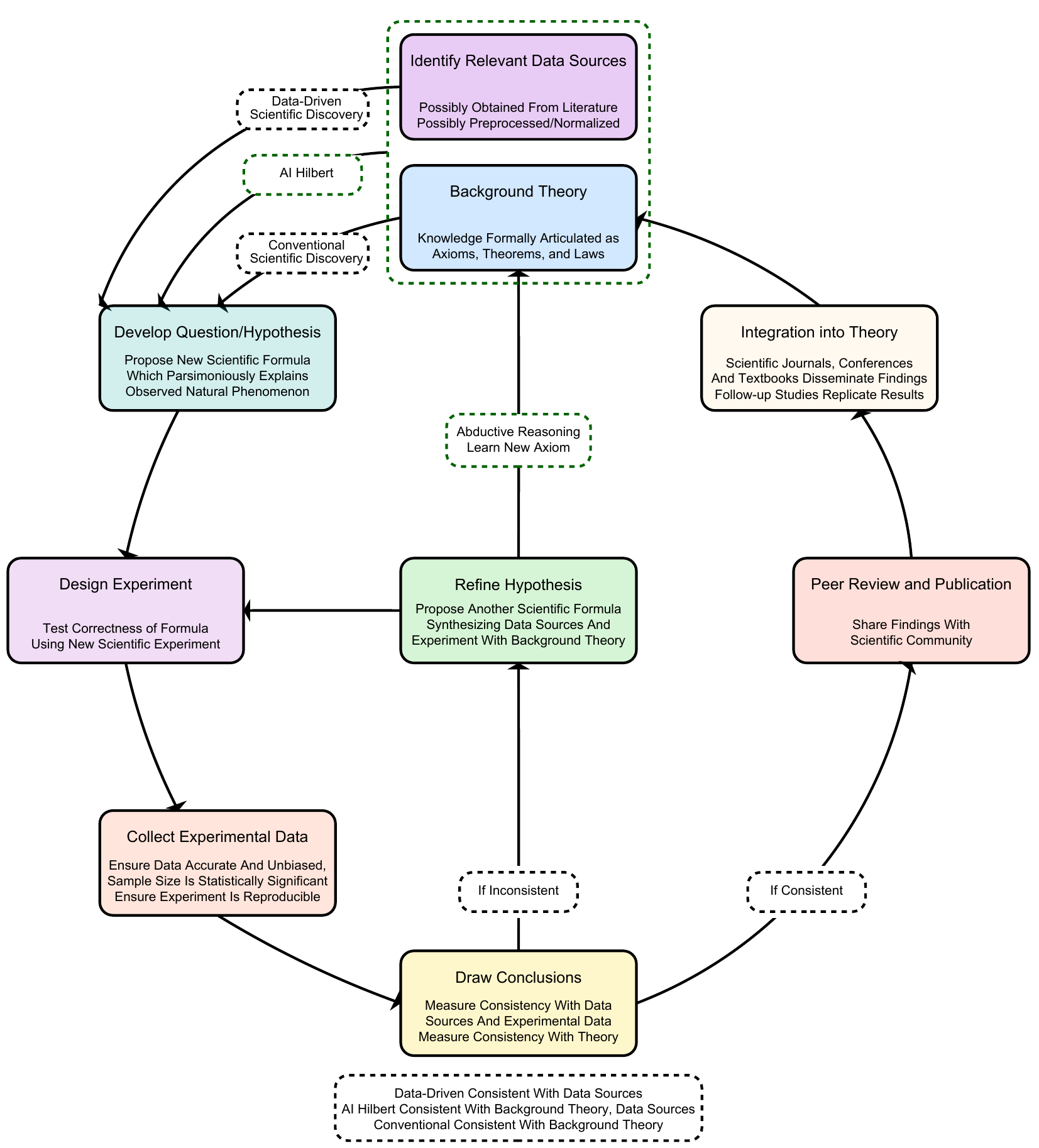}
    \caption{\textbf{The scientific method with scientific discoveries made via classical methods, data-driven methods, or AI-Hilbert.} {AI-Hilbert} proposes scientific laws consistent with a body of background theory formally articulated as polynomial equalities, inequalities, and relevant data sources. This likely allows scientific discoveries to be made using fewer data points than state-of-the-art approaches, and for missing scientific axioms to be deduced via abductive reasoning as part of the scientific discovery process. On the other hand, existing approaches to scientific discovery propose laws that may be inconsistent with either background theory or existing data sources.}
    \label{fig:scientificmethodupdated}
\end{figure}

To contrast our approach with existing approaches to scientific discovery, Figure~\ref{fig:scientificmethodupdated} depicts a stylized version of the scientific method. In this version, new laws of nature are proposed from background theory (which may be written down by humans, automatically extracted from existing literature, or {\color{black}even} generated using AI {\color{black}\cite{jumper2021highly}}) and experimental data, using classical 
discovery techniques, data-driven techniques, or \verb|AI-Hilbert|. Observe that data-driven discoveries may be inconsistent with background theory, and discoveries via classical methods may not be consistent with relevant data sources, while discoveries made via \verb|AI-Hilbert| are consistent with background theory and relevant data sources. This suggests that \verb|AI-Hilbert| could be a first step toward 
discovery frameworks that are less likely to make false discoveries. Moreover, as mentioned in the introduction, \verb|AI-Hilbert| uses background theory to restrict the effective dimension of the set of possible scientific laws, and, therefore, likely requires less 
data to make scientific discoveries than purely data-driven approaches.

\subsection{Background and Notation}\label{ssec:notation}
The notation is mostly standard to the {\color{black}semidefinite and} polynomial optimization literature. We let non-boldface characters such as $b$ denote scalars, lowercase bold-faced characters such as $\bm{x}$ denote vectors, uppercase bold-faced characters such as $\bm{A}$ denote matrices, and calligraphic uppercase characters such as $\mathcal{Z}$ denote sets. We let $[n]$ denote the set of indices $\{1, \dots, n\}$. We let $\bm{e}$ denote the vector of ones, $\bm{0}$ denote the vector of all zeros, and $\mathbb{I}$ denote the identity matrix. We let $\Vert \bm{x} \Vert_p$ denote the $p$-norm of a vector $\bm{x}$ for $p \geq 1$. We let $\mathbb{R}$ denote the real numbers, $\mathcal{S}^n$ denote the cone of $n \times n$ symmetric matrices, and $\mathcal{S}^n_+$ denote the cone of $n \times n$ positive semidefinite matrices. 

We also use some notations specific to the sum-of-squares (SOS) optimization literature; see \cite{cox2013ideals} for an introduction to computational algebraic geometry and \cite{blekherman2012semidefinite} for a general theory of sum-of-squares and convex algebraic optimization. Specifically, we let $\mathbb{R}[\bm{x}]_{n, 2d}$ denote the ring of real polynomials in the $n$-tuple of variables $\bm{x} \in \mathbb{R}^n$ of degree $2d$, $P_{n,2d}:=\{p \in \mathbb{R}[\bm{x}]_{n, 2d}: p(\bm{x}) \geq 0\quad \forall \bm{x} \in \mathbb{R}^n\}$ denote the 
convex cone of non-negative polynomials in $n$ variables of degree $2d$, and 
$$\Sigma[\bm{x}]_{n,2d}:=\left\{p(\bm{x}): \ \exists q_i, \ldots, q_m \in \mathbb{R}[\bm{x}]_{n, d}, \ p(\bm{x})=\sum_{i=1}^m q_i^2(\bm{x})\right\}$$ denote the cone of sum-of-squares polynomials in $n$ variables of degree $2d$, which can be optimized over via $n+d \choose d$ dimensional semidefinite matrices \citep[c.f.][]{parrilo2003semidefinite} using interior point methods \citep[][]{nesterov1994interior}. Note that $\Sigma[\bm{x}]_{n,2d} \subseteq P_{n,2d}$, and the inclusion is strict unless $n \leq 2$, $2d \leq 2$ or $n=3, 2d=4$ \citep[][]{hilbert1888darstellung}. Nonetheless, $\Sigma[\bm{x}]_{n,2d}$ provides a high-quality approximation of $P_{n,2d}$, since each non-negative polynomial can be approximated (in the $\ell_1$ norm of its coefficient vector) to any desired accuracy $\epsilon>0$ by a sequence of sum-of-squares \cite{lasserre2007sum}. If the maximum degree $d$ is unknown, we suppress the dependence on $d$ in our notation.

To define a notion of distance between polynomials, we also use several functional norms. Let $\Vert \cdot \Vert_p$ denote the $\ell_p$ norm of a vector. Let $\bm{\mu} \in \mathbb{N}^n$ be the vector $(\mu_1, \ldots, \mu_n)$ and $\bm{x}^{\bm{\mu}}$ stand for the monomial $x_1^{\mu_1}\ldots x_n^{\mu_n}$. Then, for a polynomial $q \in \mathbb{R}_{n,2d}[\bm{x}]$ with the decomposition $q(\bm{x})=\sum_{\bm{\mu}\in \mathbb{N}^n: \Vert \bm{\mu}\Vert_1 \leq 2d } a_{\bm{\mu}}\bm{x}^{\bm{\mu}}$, we let the notation $\Vert \bm{q}\Vert_p=\left(\sum_{\bm{\mu}\in \mathbb{N}^n: \Vert\bm{\mu}\Vert_1 \leq 2d}\alpha_{\bm{\mu}}^p\right)^{1/p}$ denote the coefficient norm of the polynomial, 

Finally, to derive new laws of nature from existing ones, we repeatedly invoke a fundamental result from real algebraic geometry called the Positivstellensatz \citep[see, e.g.,][]{stengle1974nullstellensatz}. Various versions of the Positivstellensatz exist, with stronger versions holding under stronger assumptions \citep[see][for a review]{laurent2009sums}, and any reasonable version being a viable candidate for our approach. For simplicity, we invoke a compact version due to \cite{putinar1993positive}, which holds under some relatively mild assumptions but nonetheless lends itself to relatively tractable optimization problems:

\begin{theorem}[Putinar's Positivstellensatz \cite{putinar1993positive}, see also Theorem 5.1 of \cite{parrilo2003semidefinite}]\label{thm:psatz}
Consider the basic (semi)algebraic sets
\begin{align}
\mathcal{G} := \left\{\bm{x} \in \mathbb{R}^n: \ g_1(\bm{x}) \ge 0, \ldots, g_k(\bm{x}) \ge 0\right\}, \label{ineq-axioms}\\
\mathcal{H} := \left\{\bm{x} \in \mathbb{R}^n: \ h_1(\bm{x}) =0, \ldots h_{\color{black}l} (\bm{x})=0\right\},\label{eq-axioms}
\end{align}
where $g_i, h_j \in \mathbb{R}[x]_n$, and $\mathcal{G}$ satisfies the Archimedean property\footnote{This assumption is stronger than the compactness assumption on $\mathcal{G}$ found 
in the Positivstellensatz of \cite{schmudgen1991moment}, but is typically not restrictive in practice, as one could assume that $g_1(x) = R - \sum_{i=1}^n x_i^2$ for some constant $R$. Moreover, it is arguably more tractable-it avoids the need to explicitly consider products of the form $g_i g_j$ in the decomposition, although we may require SOS polynomials of a higher degree to generate a valid certificate.} {\color{black}\citep[see also][Chap. 6.4.4]{blekherman2012semidefinite}}, i.e., there exists an $R>0$ and $\alpha_0, \ldots \alpha_k \in \Sigma[\bm{x}]_n$ such that $R - \sum_{i=1}^n x_i^2 = \alpha_0(\bm{x}) + \sum_{i=1}^k \alpha_i(\bm{x}) g_i(\bm{x})$. 

Then, for any $f \in \mathbb{R}[x]_{n, 2d}$, the implication
$$\bm{x} \in \mathcal{G} \cap \mathcal{H}   \implies f(\bm{x}) \ge 0$$ holds if and only if there exist SOS polynomials $\alpha_0, \ldots, \alpha_k \in \Sigma[\bm{x}]_{n, 2d}$, and real polynomials $\beta_1, \ldots, \beta_l \in \mathbb{R}[\bm{x}]_{n, 2d}$ such that
\begin{align}
f(x) = \alpha_0(x) + \sum_{i=1}^k \alpha_i(\bm{x}) g_i(\bm{x})+\sum_{j=1}^{\color{black}l} \beta_j(\bm{x})h_j(\bm{x}).
\end{align}
\end{theorem}

{\color{black}Note that strict polynomial inequalities of the form $h_i(x)>0$ can be modeled by introducing an auxiliary variable $\tau$ and requiring that $h_i(x) \tau^2-1=0$, and thus our focus on non-strict inequalities in Theorem \ref{thm:psatz} is without loss of generality \citep[see also][]{blekherman2012semidefinite}.}

Remarkably, the Positivstellensatz implies that if we set the degree of $\alpha_i$s to be zero, then polynomial laws consistent with a set of equality-constrained polynomials can be searched over via linear optimization. Indeed, this subset of laws is sufficiently expressive that, as we demonstrate in our numerical results, it allows us to recover Kepler's third law and Einstein's dilation law axiomatically. Moreover, the set of polynomial natural laws consistent with polynomial (in)equalities can be searched via semidefinite or sum-of-squares optimization.

We close this section by remarking that one could develop an alternative version of the Positivstellensatz with only inequality constraints, by expressing each equality via two inequalities. However, this increases the number of decision variables in the optimization problems generated by the Positivstellensatz and solved in this paper, and thus decreases the tractability of these optimization problems; see also \cite{blekherman2012semidefinite}. Accordingly, we treat equality and inequality constraints separately 
throughout.

\subsection{Structure}
The rest of the paper is organized as follows: in Section \ref{sec:application}, we describe \verb|AI-Hilbert|, the scientific discovery system proposed in this paper. 
In Section \ref{sec:experiments}, we argue that it presents an exciting new approach to scientific discovery, by demonstrating that it can rediscover the Hagen-Poiseuille Equation, Einstein's Relativistic Time Dilation Law, Kepler's Third Law, the Radiated Gravitational Wave Power Equation, and the Bell Inequalities {\color{black}(additionally, a detailed comparison of \verb|AI-Hilbert| with state-of-the-art approaches in Appendix~\ref{append:sota_comparison} and examples of scientific discovery via \verb|AI-Hilbert| in Appendix \ref{append:additionalexamples}).} In Section \ref{sec:summary}, we summarize our conclusions and discuss the limitations of and 
opportunities arising from this work.

\section{Discovering Scientific Formulae Via Polynomial Optimization} 
\label{sec:application}

In this section, we formally introduce \verb|AI-Hilbert|. 
First, {\color{black}in Section 
\ref{ssec:overview}, we provide an overview of our approach. 
Next,} in Section \ref{ssec:overallproblem}, we formalize our approach as a polynomial optimization problem. 
{\color{black}Further, } in Section \ref{ssec:distance}, we define a new notion of the distance between a polynomial and a (possibly inconsistent or incomplete) set of polynomial background knowledge axioms. 
{\color{black}Subsequently, in Section \ref{ssec:amountofdataneeded}, we discuss the role of background theory on the amount of data required to discover scientific laws, by revisiting some well-studied examples from the machine learning and statistics literature. }
{\color{black}Finally}, in Section \ref{ssec:specialcases}, we specialize our approach to problem settings where a scientist has access to a complete background theory and no experimental data.

\subsection{{Method Overview}}\label{ssec:overview}

\color{black} 

Our scientific discovery method (\verb|AI-Hilbert|) aims to discover an unknown polynomial formula $q(\cdot) \in \mathbb{R}[x]$ which describes a physical phenomenon, and is both consistent with a background theory of polynomial equalities and inequalities $\mathcal{B}$ (a set of axioms) and a collection of experimental data $\mathcal{D}$ (defined below). We provide a high-level overview of \verb|AI-Hilbert| in Figure~\ref{fig:system} and summarize our procedure in Algorithm \ref{alg:aiHilbert}. 
The inputs to \verb|AI-Hilbert| are a four-tuple $(\mathcal{B}, \mathcal{D}, \mathcal{C}(\Lambda), d^c)$, where:

\begin{figure}[b!]
    \centering
    \includegraphics[width=\linewidth]{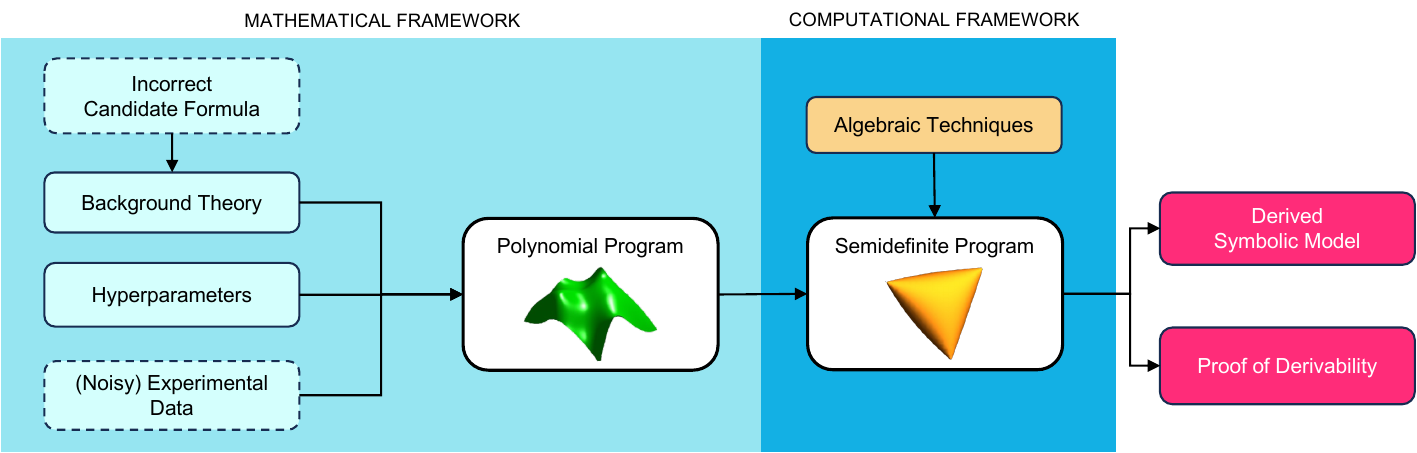}
    \caption{\textbf{Schematic illustration of AI Hilbert and its components.} \color{black} Using background knowledge 
    encoded as multivariate polynomials, experimental data, and 
    hyperparameters (e.g., a sparsity constraint on the background theory) to control our model's complexity, we formulate scientific discovery as a polynomial optimization problem, reformulate it as a semidefinite optimization problem, and solve it to obtain both a symbolic model and its formal derivation. Dashed boxes correspond to optional components. An example is introducing an incorrect candidate formula as a new axiom in the background theory. 
    }
    \label{fig:system}
\end{figure}

\begin{itemize}[leftmargin=*]
    \item $\mathcal{B}$ denotes the relevant background theory, expressed as a collection of axioms, in the scientific discovery setting, i.e., the polynomial laws relevant for discovering $q$. It is the union of the inequalities $\{g_1(\bm{x}) \geq 0 , \ldots, g_k(\bm{x}) \geq 0\}$ defining $\mathcal{G}$ and the equalities $\{h_1(\bm{x}) = 0 , \ldots, h_l(\bm{x}) = 0\}$ defining $ \mathcal{H}$ -- where $\mathcal{G}$ and $\mathcal{H}$ are as in (\ref{ineq-axioms}) and (\ref{eq-axioms}), respectively.
    $\mathcal{B}$ is defined over $n$ variables $x_1, \ldots, x_n$. However, only $t$ of these $n$ variables can be measured and are directly relevant for explaining the observed phenomenon.  {\color{black}In particular, we let $x_1$ denote the dependent variable. }
    The remaining $n-t$ variables appear in the background theory but are not directly observable\footnote{\color{black}One might also exclude measurements for certain variables $x_j$ if they are trivially connected to other measurable variables. For instance, the period of rotation can trivially be obtained from the frequency of rotation, and it is better to avoid having both in the list of measured variables in our system. We treat known quantities such as $c$ - the speed of light - as measurable entities.}.
    The background theory $\mathcal{B}$ is defined as {\it complete} if it contains all the axioms necessary to formally prove the target formula, {\it incomplete} otherwise. Moreover, $\mathcal{B}$ is called {\it inconsistent} if it contains axioms that contradict each other, {\it consistent} otherwise (we define these terms more rigorously in Section \ref{ssec:distance}). A special case of inconsistency is when a formula that incorrectly describes the studied phenomenon is added to a consistent background theory (see Section~\ref{ssec:kepler}).
\item $\mathcal{D}:=\{\bm{\bar x}_i\}_{i \in [m]}$ denotes a collection of data points, or measurements of an observed physical phenomenon, which may be contaminated by noise, e.g., from measurement error. We assume that $\bm{\bar x}_i \in \mathbb{R}^n$ and ${\bm{\bar x}_{i,j}} = 0$ for $j \geq t+1$,  i.e., the value of ${\bm{\bar x}_{i,j}}$ -- the $j$th entry of $\bar{\bm{x}_i}$ -- is set to zero for all variables $j$ that cannot or should not be measured.
\item $\mathcal{C}$ denotes a set of constraints and bounds which depend on a set of hyper-parameters $\Lambda$. Specifically, we consider a global bound on the degree of the polynomial $q$; a vector $\bm{d}$ restricting individual variable degrees in $q$; a hyperparameter $\lambda$ that models the fidelity to background theory and data; and a bound over the number of axioms that should be included in a formula derivation.
\item $d^c(\cdot, \mathcal{G}\cap \mathcal{H})$ denotes a distance function that defines the distance from an arbitrary polynomial to the background theory. We formally define $d^c$ in Section \ref{ssec:distance}.
\end{itemize}

Algorithm \ref{alg:aiHilbert} provides a high-level description of \verb|AI-Hilbert|. The procedure first combines the background theory $\mathcal{B}$ and data $\mathcal{D}$ to generate a polynomial optimization problem $\mathbbmss{Pr}$ which targets a specific concept identified by a dependent - or target - variable, included in the set of observable entities that can be measured in the environment ($x_1, \dots, x_t$). This is achieved by leveraging the distance $d^c$ (formally defined in Section \ref{ssec:distance}) and integrating the bounds and constraints $\mathcal{C}$ (with their hyperparameters $\Lambda$) via the \verb|PolyJuMP.jl| \verb|Julia| package \citep{lubin2023jump}. This corresponds to the \verb|Formulate| step of Algorithm~\ref{alg:aiHilbert}, which we formalize in Section \ref{ssec:overallproblem}.

\begin{algorithm}[h!]
\caption{AI Hilbert for Scientific Discovery}\label{alg:aiHilbert}
    \begin{algorithmic}[1]
        \REQUIRE $(\mathcal{B}, \mathcal{D}, \mathcal{C}(\Lambda), d^c) $
        \STATE $\mathbbmss{Pr} \leftarrow$ \verb|Formulate|$(\mathcal{B}, \mathcal{D}, \mathcal{C}(\Lambda), d^c)$
        \STATE $\mathbbmss{Pr^{sd}} \leftarrow $ \verb|Reduce|$(\mathbbmss{Pr})$
        \STATE $q(\bm{x}) \leftarrow $ \verb|Solve|$(\mathbbmss{Pr^{sd}})$
        \ENSURE $q(\bm{x})= 0$ 
        \ENSURE $\bm{\alpha}$, $\bm{\beta}$
    \end{algorithmic}
\end{algorithm}

\verb|AI-Hilbert| then reformulates the problem $\mathbbmss{Pr}$ as a 
semidefinite (or linear 
if no inequalities are present in the background theory) optimization problem $\mathbbmss{Pr^{sd}}$, by leveraging standard techniques from 
sum-of-squares optimization 
that are now integrated within the \verb|SumOfSquares.jl| and \verb|PolyJuMP.jl| \verb|Julia| packages, as discussed in Section~\ref{ssec:notation}. This corresponds to the \verb|Reduce| step of Algorithm~\ref{alg:aiHilbert}.

Next, \verb|AI-Hilbert| solves $\mathbbmss{Pr^{sd}}$ using a mixed-integer conic optimization solver such as \verb|Gurobi| \cite{achterberg2019s} or \verb|Mosek| \cite{andersen2000mosek}. This corresponds to the \verb|Solve| step of Algorithm~\ref{alg:aiHilbert}.

\verb|AI-Hilbert| then outputs a candidate formula of the form $q(\bm{x}) = 0$ where the only monomials with nonzero coefficients are those that only contain the variables $x_1, \ldots, x_t$, the (independent and dependent) variables that are observed in the environment. The background theory may contain additional variables $x_{t+1}, \ldots, x_n$ that are not observed in the environment and that will not appear in the derived law.  This is because the axioms in the background theory are not 
constraints on the functional form of the target polynomial, but rather 
general scientific laws describing the environment, often not including any of the quantities/variables observed in the data.

Finally, \verb|AI-Hilbert| returns multipliers $\{{\alpha}_i\}_{i=1}^k, \{\beta_j\}_{j=1}^l$ such that  $$q(\bm{x})=\alpha_0(\bm{x})+\sum_{i=1}^k \bm{\alpha}_i(\bm{x})g_i(\bm{x})+\sum_{j=1}^l \bm{\beta}_j(\bm{x})h_j(\bm{x})$$ if $d^c(q, \mathcal{G}\cap\mathcal{H})=0$, which is a certificate of the fact that $q$ is derivable from the background theory. If $d^c>0$, then \verb|AI-Hilbert| returns a certificate that $q$ is approximately derivable from the background theory, and $q$ is approximately equal to $\alpha_0(\bm{x})+\sum_{i=1}^k \bm{\alpha}_i(\bm{x})g_i(\bm{x})+\sum_{j=1}^l \bm{\beta}_j(\bm{x})h_j(\bm{x})$.



\color{black}

\subsection{Overall Problem Setting}\label{ssec:overallproblem}
\verb|AI-Hilbert| aims to discover an unknown polynomial model $q(\bm{x}) = 0$, which contains one or more dependent variables raised to some power within the expression (to avoid the trivial solution $q \equiv 0$), is approximately consistent with our axioms $\mathcal{G}$ and $\mathcal{H}$ —meaning $d^c$ is small, and explains our experimental data well, meaning $q(\bm{\bar x}_i)$ is small for each data point $i$
, and is of low complexity. 

Let $x_1, \ldots, x_t$ denote the measurable variables, and let $x_1$ denote the dependent variable which we would like to ensure appears in our scientific law. Let $\Omega = \{\bm{\mu} \in \mathbb{N}^n: \Vert \bm{\mu} \Vert_1 \leq 2d\}$. Let the discovered polynomial expression be
\[ q(x) = \sum_{\bm{\mu} \in \Omega} a_{\bm{\mu}}x^{\bm\mu},\]
{\color{black} where $2d$ is a bound on the maximum allowable degree of $q$}.  We formulate the following polynomial optimization problem:

{\color{black}
\begin{align}
\min_{q \in \mathbb{R}_{n,2d}} \quad & \sum_{\bm{\bar x}_i \in \mathcal{D}} |q(\bar{\bm{x}_i})| +\lambda \cdot d^c(q, \mathcal{G} \cap \mathcal{H})\label{prob:overal}\\
\text{s.t.} \quad & \sum_{\bm{\mu} \in \Omega: \bm{\mu}_1 \geq 1} a_{\bm{\mu}}=1, \nonumber\\
& a_{\bm{\mu}}=0  \ \ \forall \bm{\mu} \in \Omega: \sum_{j=t+1}^n \bm{\mu}_j \geq 1,\nonumber 
\end{align}
}

where $d^c$, {\color{black}the distance between $q$ and the background theory,} is the optimal value of an inner minimization problem {\color{black}we define in Section \ref{ssec:distance}}, $\lambda>0$ is a hyperparameter that balances the relative importance of model fidelity to the data against model fidelity to a set of axioms, the first constraint ensures that $x_1$, our dependent variable, appears in $q$, the second constraint ensures that we do not include any unmeasured variables.
 In certain problem settings, we constrain $d^c=0$, rather than penalizing the size of $d^c$ in the objective.

Note that the formulation of the first constraint 
controls the complexity of the scientific discovery problem via the degree of the Positivstellensatz certificate: a smaller bound on the 
allowable degree in the certificate yields a more tractable optimization problem but a less expressive family of certificates to search over, which ultimately entails a trade-off that needs to be made by the user. Indeed, this trade-off has been formally characterized by Lasserre~\cite{lasserre2007sum}, who showed that every non-negative polynomial is approximable to any desired accuracy by a sequence of sum-of-squares polynomials, with a trade-off between the degree of the SOS polynomial and the quality of the approximation.


After solving Problem \eqref{prob:overal}, one of two possibilities occurs. Either the distance between $q$ and our background information is $0$, or the Positivstellensatz provides a non-zero polynomial 
\begin{align*}\color{black}
r(\bm{x}):=q(\bm{x}) - \alpha_0(\bm{x}) - \sum_{i=1}^k \alpha_i(\bm{x}) g_i(\bm{x})-\sum_{j=1}^{\color{black}l} \beta_j(\bm{x}) h_j(\bm{x})
\end{align*}
which defines the discrepancy between our derived physical law and its projection onto our background information. In this sense, solving Problem \eqref{prob:overal} also provides information about the inverse problem of identifying a complete set of axioms that explain $q$. In either case, it follows from the Positivstellensatz (Theorem \ref{thm:psatz}) that solving Problem \eqref{prob:overal} for different hyperparameter values and different bounds on the degree of $q$ eventually yields polynomials that explain the experimental data well and are approximately derivable from background theory.

We close this section with two remarks on the generality and complexity of \verb|AI-Hilbert|. 

\paragraph{Implicit and Explicit Symbolic Discovery:} 
Most prior work \citep[e.g., ][]{curmei2020shape, cornelio2021ai,schmidt2009symbolic}
 aims to identify an unknown symbolic model {\color{black}$f \in \mathbb{R}[x]_{n,2d}$} of the form $y_i = f(\bm{x}_i)$ for a set of independent variables of interest $\bm{x}_i \in \mathbb{R}^n$ and a dependent variable $y_i \in \mathbb{R}$, while
 \verb|AI-Hilbert|
 searches for an implicit polynomial function $q$ which links the dependent and independent variables. We do this for two reasons. First, many scientific formulae of practical interest admit implicit representations as polynomials, but explicit representations of the dependent variable as a polynomial function of the independent variables are not possible  \citep[c.f.][]{ahmadi2019polynomial}. 
For instance, Kepler's third law of planetary motion has this property {\color{black}(see Section \ref{ssec:kepler})}. 
Second, as proven by Artin~\cite{artin1927zerlegung} to partially resolve Hilbert's $17$th problem \citep[c.f.][]{hilbert1888darstellung}, an arbitrary non-negative polynomial can be represented as a sum of squares of rational functions. Therefore, by multiplying by the denominator in Artin's representation~\cite{artin1927zerlegung}, 
implicit representations of natural laws become a viable and computationally affordable search space.

{\color{black}We remark that the implicit representation of scientific laws as polynomials where $q(\bm{x})=0$ introduces some degeneracy in the set of optimal polynomials derivable from \eqref{prob:overal}, particularly in the presence of a correct yet overdetermined background theory. For instance, in the derivation of Kepler's Law of Planetary Motion in Section \ref{ssec:kepler}, we eventually derive the polynomial $m_1 m_2 G p^2=m_1 d_1 d_2^2+m_2 d_1^2 d_2+2 m_2 d_1 d_2^2$. Since we have the axiom that $m_1 d_1=m_2 d_2$, we could instead derive the (equivalent) formula $m_1 m_2 G p^2=(m_1+m_2)d_1 d_2 (d_1+d_2)$. To partly break this degeneracy, we propose to either constrain the degree of the proof certificate and gradually increase it (as is done in \eqref{prob:overal}) or, (equivalently in a Lagrangian sense) include a term modeling the complexity of our derived polynomial (e.g., $\Vert q\Vert_1$, the $L_1$-coefficient norm of $q$) in the objective.}

\paragraph{Complexity of Scientific Discovery:} Observe that, if the degree of our new scientific law $q$ is fixed and the degree of the polynomial multipliers in the definition in $d^c$ is also fixed, then Problem \eqref{prob:overal} can be solved in polynomial time\footnote{Under the real number complexity model, and under the bit number complexity model under some mild regularity conditions on the semidefinite optimization problems that arise from our sum-of-squares optimization problems. Note that, under the bit complexity model, semidefinite optimization problems cannot always be solved in polynomial time due to the existence of ill-behaved semidefinite problems where all feasible solutions are of doubly exponential size. We refer to Ramana~\cite{ramana1997exact} or Laurent and Rendl~\cite{laurent2005semidefinite} for a complete characterization of the complexity of semidefinite optimization.} with a consistent set of axioms (resp. nondeterministic polynomial time with an inconsistent set of axioms). This is because solving Problem \eqref{prob:overal} with a fixed degree and a consistent set of axioms corresponds to solving a semidefinite optimization problem of a polynomial size, which can be solved in polynomial time (assuming that a constraint qualification such as Slater's condition holds) \cite{nesterov1994interior}. Moreover, although solving Problem \eqref{prob:overal} with a fixed degree and an inconsistent set of axioms corresponds to solving a mixed-integer semidefinite optimization problem, which is NP-hard, recent evidence \cite{dey2021branch} shows that integer optimization problems can be solved in polynomial time with high probability. This suggests that Problem \eqref{prob:overal} may also be solvable in polynomial time with high probability. However, if the degree of $q$ is unbounded then, to the best of our knowledge, no existing algorithm solves Problem \eqref{prob:overal} in polynomial time. This explains why searching for scientific laws of a fixed degree and iteratively increasing the degree of the polynomial laws searched over, in accordance with Occam's Razor, is a key aspect of our approach.

\subsection{Distance to Background Theory and Model Complexity}\label{ssec:distance}
{\color{black}Scientists often start with experimental measurements and a set of polynomial equalities and inequalities (axioms) which they believe to be true.} 
From these axioms and measurements, they aim to deduce a new law, explaining their data, which includes one or more dependent variables and excludes certain variables.
The simplest case of scientific discovery involves a consistent and correct set of axioms that fully characterize the problem. In this case, the 
Positivstellensatz {\color{black}(Theorem \ref{thm:psatz}) facilitates} 
the discovery of new scientific laws via deductive reasoning, without using any experimental data, as we argue in Section \ref{ssec:specialcases}. Indeed, under an Archimedean assumption, the set of all valid scientific laws corresponds precisely to the preprime (see~\cite{cox2013ideals} for a definition) generated by our axioms~\cite{putinar1993positive}, and searching for the simplest polynomial version of a law which features a given dependent variable corresponds to solving an easy linear or semidefinite feasibility problem. 

It is not uncommon to have a set of axioms that is inconsistent (meaning that there are no values of $\bm{x} \in \mathbb{R}^n$ that satisfy all laws simultaneously), or incomplete (meaning the axioms do not `span' the space of all derivable polynomials; we provide a formal definition later in this section). Therefore, we require a notion of a distance between a body of background theory (which, in our case, consists of a set of polynomial equalities and inequalities) and a polynomial. We now establish this definition, treating the inconsistent and incomplete cases separately; {\color{black}note that the inconsistent and incomplete case may be treated via the inconsistent case}. We remark that \citep[]{blekherman2012semidefinite, zhao2023hausdorff} propose related notions of the distance between (a) a point and a variety defined by a set of equality constraints, and (b) the distance between two semialgebraic sets via their Hausdorff distance. However, to our knowledge, the distance metrics 
in this paper have not previously been proposed.

\paragraph{Incomplete Case:} Suppose $\mathcal{B}$ is a background theory  (consisting of equalities and inequalities in $\mathcal{G}$ and $\mathcal{H}$) 
, where $\mathcal{G}$ satisfies the previously defined Archimedean property (see Theorem \ref{thm:psatz}) with constant $R$, and the axioms are not inconsistent, meaning that $\mathcal{G} \cap \mathcal{H} \neq \emptyset$. Then, a natural notion of distance is the $\ell_2$ coefficient distance $d^c$ between $q$ and $\mathcal{G} \cap \mathcal{H}$, which is given by: 
$$d^c (q, \mathcal{G} \cap \mathcal{H})  := \min_{\substack{\alpha_0, \ldots, \alpha_k \in \Sigma_{n,2d}[\bm{x}], \\\beta_1, \ldots, \beta_{\color{black}l} \in \mathbb{R}_{n,2d}}}  \left\Vert q- \alpha_0 - \sum_{i=1}^k \alpha_i g_i-\sum_{j=1}^{\color{black}l} \beta_j h_j\right\Vert_2. $$
It follows directly from Putinar's Positivstellensatz that $d(q, \mathcal{G} \cap \mathcal{H})=0$ if and only if $q$ is derivable from $\mathcal{B}$. We remark that this distance has a geometric interpretation as the distance between a polynomial $q$ and its projection onto the algebraic variety generated by $\mathcal{G} \cap \mathcal{H}$. Moreover, by norm equivalence, this is equivalent to the Hausdorff distance \cite{zhao2023hausdorff} between $q$ and $\mathcal{G} \cap \mathcal{H}$.

With the above definition of $d^c$, and the fact that $\mathcal{G} \cap \mathcal{H}\neq \emptyset$, we say that $\mathcal{G} \cap \mathcal{H}$ is an incomplete set of axioms if there does not exist a polynomial $p$ with a non-zero coefficient on at least one monomial involving a dependent variable, such that $d^c(q, \mathcal{G} \cap \mathcal{H})=0$. 

\paragraph{Inconsistent Case:} Suppose $\mathcal{B}$ is an inconsistent background theory i.e, 
$\mathcal{G} \cap \mathcal{H}=\emptyset$. 
Then, a natural approach to scientific discovery is to assume that a subset of the axioms are valid laws, while the remaining axioms are scientifically invalid (or invalid in a specific context, e.g., micro vs. macro-scale).
In line with the sparse regression literature \citep[c.f.][]{ bertsimas2016best} and related work on discovering nonlinear dynamics \citep{bertsimas2023learning, liu2024okridge}, we assume that scientific discoveries can be made using at most $k$ correct scientific laws and define the distance between the scientific law and the problem data via a best subset selection problem. Specifically, we introduce binary variables $z_i$ {\color{black}and $y_j$} 
to denote whether the $i$th {\color{black}and $j$-th laws are consistent}, and require that $\alpha_i=0$ if $z_i=0$ {\color{black}and $\beta_j=0$ if $y_j=0$ and $\sum_i z_i+\sum_j y_j \leq \tau$ for a sparsity budget $\tau$}. Furthermore, we allow a non-zero $\ell_2$ distance between the scientific law $f$ and the reduced background theory, but penalize this distance in the objective. This gives the following notion of distance between a scientific law $q$ and the space $\mathcal{G} \cap \mathcal{H}$:
\begin{align*}
    d^c (q, \mathcal \mathcal{G} \cap \mathcal{H})  := \min_{} \quad &  \left\Vert q- \alpha_0 - \sum_{i=1}^{\color{black}k} \alpha_i g_i-\sum_{j=1}^{\color{black}l} \beta_j h_j \right\Vert_2, \\
    \text{s.t.} \quad & \alpha_i=0 \ \text{if} \ z_i=0, \forall i \in \{0, \ldots, {\color{black}k}\},\\
    & \beta_j=0 \ \text{if} \ y_j=0, \ \forall j \in \{1, \ldots {\color{black}l}\},\\
    & \sum_{i=0}^{\color{black}k} z_i+\sum_{j=1}^{\color{black}l} y_{\color{black}j} \leq \tau,\\
    & z_0 \ldots z_{{\color{black}k}} \in \{0, 1\}, \ y_1, \ldots y_{\color{black}l} \in \{0, 1\},\\
    & \alpha_0, \ldots, \alpha_{\color{black}k} \in \Sigma_{n,2d}[\bm{x}], 
    \ \beta_1, \ldots, \beta_{\color{black}l} \in {\color{black}\mathbb{R}[x]_{n,2d}}.
\end{align*}

It follows directly from the Positivstellensatz that $d=0$ if and only if $q$ can be derived from $\mathcal{B}$. If $\tau=m+l$, then we certainly have $d^c=0$, since the overall system of polynomials is inconsistent and the sum-of-squares proof system can deduce that $``-1 \geq 0$'' from inconsistent proof systems, from which it can claim a distance of $0$. However, by treating $\tau$ as a hyper-parameter and including the quality of the law on experimental data as part of the optimization problem (see Section \ref{ssec:overallproblem}), scientific discoveries can be made from inconsistent axioms by incentivizing solvers to set $z_i=0$ for inconsistent axioms $i$. {\color{black}Alternatively, a practitioner may wish to explore the Pareto frontier of scientific discoveries that arise as we vary $\tau$, to detect how large the set of correct background knowledge is. }Provided there is a sufficiently high penalty cost on poorly explaining scientific data via the derived law, our optimization problem prefers a subset of correct axioms with a non-zero distance $d^c$ to the derived polynomial over a set of inconsistent axioms which gives a distance $d^c=0$.

\subsection{Impact of Background Theory on Amount of Data Needed to Discover Scientific Laws}\label{ssec:amountofdataneeded}
In the introduction, we suggested that providing a partially complete background theory expressible as polynomial equalities and inequalities may accelerate the scientific discovery process by decreasing the amount of data required to recover a scientific law with high probability. We now justify {\color{black}our claim in the introduction, by reviewing} examples from the machine learning literature, which may be viewed as special cases of scientific discovery, where including relevant background theory decreases the number of data points required to recover a scientific law with high probability. The first two examples involve discovery settings where the ground truth is known to be sparse, and imposing a sparsity constraint on the discovered law reduces the amount of data required to recover the law with high probability. {\color{black}Note that these examples are due to \cite{gamarnik2017high, arous2020free, candes2010matrix}. However, the similarities between them strongly suggest that similar relations between the amount of background theory introduced and the amount of data required for scientific discovery also hold in other contexts.}

\begin{example}{Sparse Linear Regression \citep{gamarnik2017high}}\\
    Consider a sparse high-dimensional regression model where we aim to recover a $\tau$-sparse regression model $\bm{\beta}^\star \in \mathbb{R}^p$ given access to $n$ noisy linear observations of the form $\bm{Y}=\bm{X}\bm{\beta}^\star+\bm{W} \in \mathbb{R}^n$, where $X_{i,j} \overset{\mathrm{iid}}{\sim} \mathcal{N}(0,1)$ and $W_i \overset{\mathrm{iid}}{\sim} \mathcal{N}(0,\sigma^2)$ for some parameter $\sigma>0$, it is known that $\bm{\beta}^\star$ is a $\tau$-sparse vector with binary coefficients, and $n \ll p$ with $p \rightarrow \infty$. Then, it is information-theoretically impossible to recover $\bm{\beta}$ if $n \leq \Theta\left(\frac{2\tau \log p}{\log\left(1+\frac{2\tau}{\sigma^2}\right)}\right)$ \cite{wang2010information}. On the other hand, for $n \geq \Theta\left(\frac{2\tau \log p}{\log\left(1+\frac{2\tau}{\sigma^2}\right)}\right)$, the (unique) optimal solution of the polynomial optimization problem (expressible as a binary problem)
    \begin{align*}
        \min_{\bm{\beta} \in \mathbb{R}^p} 
        \quad & \Vert \bm{Y}-\bm{X}\bm{\beta}\Vert_2^2 \ \text{s.t.} \ \beta_i^2=\beta_i \ \forall i \in [p], \sum_{i \in [p]}\beta_i = \tau
    \end{align*}
    is such that 
    \begin{align*}
       \frac{1}{\tau} \Vert \bm{\beta}^\star-\bm{\beta}\Vert_0 \rightarrow 0
    \end{align*}
    with high probability as $\tau \rightarrow \infty$. On the other hand, for any $\lambda>0$, the optimal solution of the popular Lasso method
    \begin{align*}
        \min_{\bm{\beta} \in \mathbb{R}^p} 
        \quad & \Vert \bm{Y}-\bm{X}\bm{\beta}\Vert_2^2 +\lambda \Vert \bm{\beta}\Vert_1
    \end{align*}
    only recovers $\bm{\beta}^\star$ with high probability when $n \geq \Theta (2\tau+\sigma^2)\log p$. We remind the reader that $$\Theta (2\tau+\sigma^2)\log p > \Theta\left(\frac{2\tau \log p}{\log\left(1+\frac{2\tau}{\sigma^2}\right)}\right).$$
\end{example}

\begin{example}{Sparse Principal Component Analysis \citep{arous2020free}}\\
Consider a sparse principal component analysis setting where we aim to recover a $\tau$-sparse binary vector $\bm{x}^\star \in \mathbb{R}^n$ given an observed matrix $\bm{Y}=\frac{\lambda}{\tau}\bm{x}^\star{\bm{x}^\star}^\top+\bm{W} $, where $\bm{W}$ is a GOE$(n)$ matrix, i.e., is a symmetric matrix with on-diagonal entries taking values $W_{i,i} \overset{\mathrm{iid}}{\sim} \mathcal{N}(0,\frac{2}{n})$ and off-diagonal entries taking values $W_{i,j} \overset{\mathrm{iid}}{\sim} \mathcal{N}(0,\frac{1}{n})$, and $\lambda>0$ is the signal-to-noise ratio. Let $\lambda, \tau$ possibly depend on $n$, and set $1 
\ll \tau \ll n$ as $n \rightarrow \infty$. Then:
\begin{itemize}
    \item Recovery of $\bm{x}^\star$ is information-theoretically impossible when $\lambda \ll \frac{\sqrt{\tau}}{\sqrt{n}}$, with high probability.
    \item The (mixed-integer representable) polynomial optimization problem
    \begin{align*}
        \max_{\bm{x} \in \mathbb{R}^n} \quad & \bm{x}^\top \bm{Y}\bm{x} \ \text{s.t.} \ x_i^2=x_i \ \forall i \in [n], \ \sum_{i \in [n]}x_i=\tau
    \end{align*}
    achieves exact recovery with high probability when $\lambda \gg \frac{\sqrt{\tau}}{\sqrt{n}}$.
    \item The diagonal thresholding algorithm of \cite{johnstone2009consistency} recovers $\bm{x}^\star$ with high probability if $\lambda \gg \frac{\tau}{\sqrt{n}}$.
    \item The vanilla PCA method, which disregards backgrounds theory encoded via a sparsity constraint and solves the polynomial optimization problem
     \begin{align*}
        \max_{\bm{x} \in \mathbb{R}^n} \quad & \bm{x}^\top \bm{Y}\bm{x} \ \text{s.t.} \ \Vert \bm{x}\Vert_2^2=\tau
    \end{align*}
    fails to recover $\bm{x}^\star$ with high probability when $\lambda >1$.
\end{itemize}
    
\end{example}

\begin{example}{Low-Rank Matrix Completion \citep{candes2010matrix}}\\
Consider a low-rank matrix completion setting where we aim to recover a fixed rank $r$ $n \times n$ matrix $\bm{A}$ given a uniform random sample of its entries $\Omega \subseteq [n] \times [n]$ of size $m$, where $\bm{A}$ satisfies the mutual incoherence property of \cite{candes2010matrix} with constant $\mu$. Then:
\begin{itemize}
    \item Recovery of $\bm{A}$ is information-theoretically impossible when $m \leq \Theta(n r \log n)$, because there are infinitely many rank-$r$ matrices that match all observed entries perfectly \cite{candes2010matrix}.
    \item The polynomial optimization problem \citep[c.f.][]{bertsimas2022mixed, bertsimas2023matrix}
    \begin{align*}
        \min_{\bm{X} \in \mathbb{R}^{n \times n}, \bm{Y} \in \mathcal{S}^n}\quad & \sum_{(i,j) \in \Omega}(X_{i,j}-A_{i,j})^2 \ \text{s.t.} \ \bm{Y}\bm{X}=\bm{X}, \bm{Y}^2=\bm{Y}, \mathrm{tr}(\bm{Y})=r
    \end{align*}
    achieves exact recovery with high probability provided $m \geq \Theta(n r \log n)$.
    \item The nuclear norm relaxation of \cite{candes2010matrix} recovers $\bm{A}$ with high probability provided $m \geq \Theta(n^{6/5}r \log n)$.
    \item The naive approach of disregarding the rank constraint and solving
    \begin{align*}
        \min_{\bm{X} \in \mathbb{R}^{n \times n}} \sum_{(i,j) \in \Omega}(X_{i,j}-A_{i,j})^2,
    \end{align*}
    which admits the solution $X_{i,j}=0$ if $(i,j) \notin \Omega$, fails to recover $\bm{A}$ with high probability provided $m < n^2$.
\end{itemize}
    
\end{example}

The above examples are admittedly more stylized than many scientific discovery settings that arise in practice. Nonetheless, they reveal that in certain circumstances, encoding relevant background provably reduces the amount of data required to recover a scientific law with high probability. This agrees with intuition: if we provide a complete background theory that can be manipulated to recover the scientific law, then, as discussed in the next section, we require no data to recover a scientific law. On the other hand, if we provide no background theory we may require a significant amount of data to recover a law. Therefore, providing relevant background theory that constrains the space of derivable scientific laws should decrease the amount of data needed to recover a scientific law with high confidence. This observation highlights the value of embedding relevant background theory within the scientific discovery process.

\subsection{Discovering Scientific Laws From Background Theory Alone}\label{ssec:specialcases}
Suppose that the background theory $\mathcal{B}$ constitutes a complete set of axioms that fully describes our physical system. Then, any polynomial which contains our dependent variable $x_1$ and is derivable from our system of axioms is a valid physical law. Therefore, we need not even collect any experimental data, and we can solve the following feasibility problem to discover a valid law (let $\Omega = \{\bm{\mu} \in \mathbb{N}^n: \Vert \bm{\mu} \Vert_1 \leq 2d\}$):

\begin{align}
\exists \quad & q(x) = \sum_{\bm{\mu} \in \Omega} a_{\bm{\mu}}x^{\bm\mu} \label{prob:existence}\\
\text{s.t.} \quad & q(\bm{x})=\alpha_0(\bm{x})+\sum_{j=1}^k \alpha_i (\bm{x}) g_i(\bm{x})+\sum_{j=1}^{\color{black}l} \beta_j  (\bm{x}) h_j(\bm{x}),\nonumber \\
\quad & \sum_{\bm{\mu} \in \Omega: \bm{\mu}_1 \geq 1} a_{\bm{\mu}}=1, \nonumber\\
& a_{\bm{\mu}}=0  \ \ \forall \bm{\mu} \in \Omega: \sum_{j=t+1}^n \bm{\mu}_j \geq 1,\nonumber \\
& \alpha_i(\bm{x}) \in \Sigma[\bm{x}]_{n,2d}, \ \beta_j(\bm{x}) \in \mathbb{R}[\bm{x}]_{n,2d},\nonumber
\end{align}
where the second and third constraints ensure that we include the dependent variable $x_1$ in our formula $q$ and rule out the trivial solution $q=0$, and exclude any solutions $q$ which contain uninteresting symbolic variables respectively.

Note that if we do not have any inequality constraints in either problem, then we may eliminate $\alpha_i$ and obtain a linear optimization problem. 

\section{Experimental Validation}\label{sec:experiments}

{In this section, we showcase AI-Hilbert's ability to recover valid scientific laws across various problem settings, including settings with incomplete or inconsistent axioms and experimental data. }

\color{black}

The rest of this section is laid out as follows: First, we provide a detailed explanation of AI-Hilbert's implementation in Section \ref{ssec:implementation}. Subsequently, we discuss the trade-off between data and theory made by \verb|AI-Hilbert| in Section \ref{ssec:tradeoffdatatheory}. 
Next, we validate \verb|AI-Hilbert| on five selected problems that highlight its different capabilities, as discussed below. \color{black}
Our main findings are as follows:

\begin{enumerate}[leftmargin=12pt]
\item In Sections~\ref{ssec:Hagen}--\ref{ssec:gravwaves}, we demonstrate that  \verb|AI-Hilbert| successfully derives valid scientific laws solely from a complete and consistent background theory. Namely, by deriving the Hagen-Poissuille Equation (Section \ref{ssec:Hagen}) and deriving the Radiated Gravitational Wave Power Equation (Section \ref{ssec:gravwaves}). We further demonstrate this capability on a suite of test instances in Appendix~\ref{append:additionalexamples}. 
\item In Sections~\ref{ssec:einstein}--\ref{ssec:kepler}, we demonstrate that \verb|AI-Hilbert| is capable of deriving a valid scientific law from an inconsistent yet complete background theory. We illustrate this capability through two scenarios:
Firstly, when two axioms in the background theory contradict each other (Section \ref{ssec:einstein}), by using measurement data to discern the correct axiom. 
Secondly, where an incorrect candidate formula is incorporated into a complete and consistent background theory (Section \ref{ssec:kepler}). In both cases, we use real-life, noisy, experimental data. 
\item In Section~\ref{ssec:kepler2.0}, we demonstrate that \verb|AI-Hilbert| successfully derives a valid scientific law from an incomplete yet consistent background theory and experimental data. Specifically, we demonstrate that \verb|AI-Hilbert| is data-efficient, in the sense that as the number of axioms supplied increases, less data is required to successfully derive a scientific law.
\item In Section~\ref{ssec:bell}, we showcase \verb|AI-Hilbert|'s ability to handle and discover inequalities, namely the Bell Inequalities. This is notable because existing scientific discovery methods, to our knowledge, cannot handle inequalities.
\item We compare AI-Hilbert with some widely used methods from the literature on a suite of test instances in Appendix~\ref{append:sota_comparison}. We demonstrate that \verb|AI-Hilbert| outperforms these methods by rediscovering more of the scientific laws in this test bed. This occurs because \verb|AI-Hilbert| integrates data and theory systematically via sum-of-squares optimization. 
\end{enumerate}
\color{black}

\color{black}

\subsection{{Implementation Details}}\label{ssec:implementation}

\color{black}
\newcommand{\monn}{\textup{mon}}

We now illustrate the lower-level implementation of \verb|AI-Hilbert| via a 
synthetic example where the axioms are assumed to be consistent and complete.  Consider a semialgebraic system in two real variables $x$ and $y$ which comprises the axioms:
\begin{eqnarray}
x^2 + y^2 - 2 =  0, \label{smallex1}\\
y - x^3  =  0, \label{smallex2}
\end{eqnarray}
where we let $h_1(x, y) = 0$ and $h_2(x,y) = 0$ denote Equations \eqref{smallex1}--\eqref{smallex2}, respectively. These axioms can be viewed as being true in a subdomain of $\mathbb{R}^2$, i.e.,  when $|x| = 1$. 

Let the set $S = \left\{(1, 1), (-1, -1)\right\}$ contain all points satisfying equations \eqref{smallex1}--\eqref{smallex2}. If
\begin{equation}\label{smallex} q(x, y) = \beta_1(x,y) h_1(x,y) + \beta_2(x,y)h_2(x,y), \end{equation} where $\beta_1, \beta_2$ are polynomials in $x, y$, then $q$ is a polynomial that vanishes on $S$. 
On the other hand, to certify that a polynomial $q(x,y)$ vanishes on $S$,  we search for a polynomial $q(x,y)$ such that $q(x,y) = g(x,y)p(x,y)$ and $q$ satisfies the expression in \eqref{smallex}.
For example, $q(x,y) = x-y$ vanishes on $S$; setting $\beta_1 = -\frac{1}{2}x, \beta_2 = -1$, we have
\begin{equation}\label{deriv-ex} q = \beta_1 h_1 + \beta_2h_2 = x-y + \frac{1}{2}(x^3 - xy^2) = (x-y)(1 + \frac{1}{2}x(x+y)). \end{equation}

Suppose we wish to find a polynomial function $q(x,y)$ that vanishes on $S$ and explains the dataset
\begin{equation*}
\bar{\bm{x}} = \left[\begin{array}{cc}
.5 &. 5 \\
-2 & -2\\
 3 & 3\\
-3 &-3
\end{array} \right]
\end{equation*}
where the columns are observed values of $x$ and $y$, respectively, and each row represents a datapoint. Note that these data observations are noiseless observations from the polynomial $f(x,y)=x-y$, which we aim to recover. Then we search for $q$ satisfying \eqref{smallex} that fits $\bar{\bm{x}}$.

We assume that $\beta_1, \beta_2, q$ are unknown polynomials that are comprised of monomials of degree at most 3, i.e., the ten monomials in the vector
\[ \monn = (1, y, y^2, y^3, x, xy, xy^2, x^2, x^2y, x^3),\]
where $\monn_1 = 1, \monn_2 = y, \ldots, \monn_{10} = x^3$.
Let
\begin{eqnarray}
    &&\beta_1(x,y) = \sum_{j=1}^{10} a_j \monn_j, \ \beta_2(x,y) = \sum_{j=1}^{10} b_j \monn_j, \mbox{ and }  q(x,y) = \sum_{j=1}^{10} c_j \monn_j. 
\end{eqnarray}
Further, let $\bm{v}$ denote the vector $(a_1, \ldots, a_{10}, b_1, \ldots, b_{10}, c_1, \ldots, c_{10})$, $\bar{\bm{x}_i}$ denote the $i$th row of $\bar{\bm{x}}$, and $q(\bar{\bm{x}_i})$ be the value of the polynomial $q$ evaluated at the point $(\bar x,\bar y) = \bar{\bm{x}_i}$.
Note that $q(\bar{\bm{x}_i}) = \sum_{j=1}^{10} c_j \monn_j(\bar{\bm{x}_i})$ 
is a linear function of the unknowns $c_1, \ldots, c_{10}$. 

If $\bar{\bm{x}_i}$ is a noiseless experimental observation, we should have $q(\bar{\bm{x}_i}) = 0$. Accordingly, we interpret any nonzero value of $|q(\bar{\bm{x}_i})|$ as the error when $q$ is evaluated at $\bar{\bm{x}_i}$, and aim to minimize this error when selecting $q$.
Equation \eqref{smallex} implies the following linear equations in the variables $a_i, b_i, c_i$, which are obtained by equating the coefficients of the monomials in Equation \eqref{smallex}:
\begin{eqnarray}
&c_1 = -2 a_1 & [\mbox{coef. of }1] \label{f1ex} \\
&c_2 = -2 a_2 + b_1 & [\mbox{coef. of }y] \label{f2ex}\\
&c_3 = a_1 - 2 a_3 + b_2 & [\mbox{coef. of }y^2] \label{f3ex}\\
&\vdots \nonumber \\
& 0 = a_3 + b_4 &[\mbox{coef. of } xy^3] \label{f11ex}
\end{eqnarray}
Let these constraints be denoted by $Av = {\bf 0}$, 
and let $m$, the number of data points, be an integer between 1 and 4. Then, we solve the linear optimization problem
\begin{eqnarray}
\min \ & 100\left(\sum_{i=1}^{m} t_i\right) + \sum_{j=1}^{10} w_j \label{prob:lphilbert}\\
\text{s.t} \ & t_i \geq q(\bar{\bm{x_i}}) & i=1, \ldots, m\nonumber\\
& t_i \geq -q(\bar{\bm{x_i}}) & i=1, \ldots, m\nonumber\\
& w_j \geq c_j & j=1,\ldots, 10\nonumber\\
& w_j \geq -c_j & j=1, \ldots, 10\nonumber\\
& Av = \mathbf{0} \nonumber\\
& \sum_{j=4}^{10} c_j = 1\nonumber\\
& v \in \mathbb{R}^{30}, t_i, w_j \geq 0 & i=1, \ldots, d, j=1, \ldots, 10.\nonumber
\end{eqnarray}

In Problem \eqref{prob:lphilbert}, the first two constraints imply that $t_i \geq |q(\bm{\bar{x}_i})|$; the third and fourth force $w_i \geq |c_i|$. The second-to-last constraint forces a monomial containing $x$ to be present in $q$ to avoid the trivial solution $q \equiv 0$. We sometimes use a different right-hand-side value to get non-fractional $c_i$ values. Thus, the optimization problem above searches for a polynomial $q$ that minimizes a weighted combination of the $L_1$-error of $q$ and the $L_1$-coefficient norm of $q$. The latter term is a {regularization} term incentivizing a sparse $q$.

If we set $m=1$ and solve \eqref{prob:lphilbert}, we obtain the (correct) function $q(x,y)=x-y$ in \eqref{prob:lphilbert}. In other words, a single data point from $\bar{\bm{x}}$ suffices to ``recover'' $x-y$ with the choice of objective in \eqref{prob:lphilbert}. 

However, let LPF be the linear optimization problem obtained by dropping the second term in the objective in \eqref{prob:lphilbert} (and not incentivizing sparsity).
If we solve LPF with $m=1$, we get $q' = 4x^3 - x^2 - y^2 - 4y+2$ as a solution. This is not a multiple of $(x-y)$ but vanishes at the points in $S$ and the first datapoint in $\mathcal{D}$. 
We need $m=2$ (and use two datapoints) before we get
\[ q' = y^3 + x^2y - 2xy^2 + 4x - 4y = (x-y)\big((x-y)^2 + 4\big).\] This illustrates the role of sparsity and regularization in reducing the amount of data required to recover a scientific law. Note that $(x-y)^2+4$ is strictly positive for all real $x,y$, and therefore $q' = 0$ on $S$ indicates that $x-y = 0$ on $S$.

Further, if we drop the constraints $Av = {\bf 0}$ and allow $q$ to be an arbitrary degree-3 polynomial in $x,y$ (and not equal to $\beta_1h_1 + \beta_2h_2$), then we cannot recover a multiple of $x-y$ until we set $m=4$ and use all datapoints in $\bar{\bm{x}}$. In the latter case, we get $q = x^3 - y^3.$ This highlights the value of background theory in restricting the space of feasible scientific laws and reducing the amount of data required to recover a scientific law.

If the only measured variable is $x$, then we would try to eliminate $y$ in the final formula. Multiplying \eqref{smallex2} by $y+x^3$ and subtracting the result from \eqref{smallex1}, we get the expression $x^2 + x^6 - 2 = 0$. This satisfies the $x$-components of points in $S$. Sometimes, there may be a unique way of eliminating variables and getting a formula on the measured variables. 

\color{black}

\subsection{{Trade off Between Data and Theory}}\label{ssec:tradeoffdatatheory}

\color{black}

There is a fundamental trade-off between the amount of background theory available and the amount of data needed for scientific discovery. Indeed, with a complete and consistent set of background theories, it is possible to perform scientific discovery without any experimental data via the Positivestellensatz (see Section \ref{ssec:notation}). On the other hand, the purely data-driven approaches to scientific discovery reviewed in Section \ref{sec:litrev} often require many noiseless or low-noise experimental observations to successfully perform scientific discovery. More generally, as the amount of relevant background theory increases, the amount of experimental data required by \verb|AI-Hilbert| to perform scientific discovery cannot increase, because increasing the amount of background theory decreases the effective VC-dimension of our scientific discovery problem \citep[see, e.g.,][]{hastie2009elements}. This can be seen in the machine learning examples discussed in Section \ref{ssec:amountofdataneeded}, where imposing a sparsity constraint (i.e., providing a relevant axiom) reduces the number of data observations needed to discover a ground truth model; see also \cite{lim2012consistency, guntuboyina2018nonparametric} for an analysis of the asymptotic consistency of shape-constrained regression. We provide further evidence of this experimentally in Section \ref{ssec:kepler2.0}.

The above observations can also be explained via real algebraic geometry: requiring that a scientific law is consistent with an axiom is equivalent to restricting the space of valid scientific laws to a subset of the space of discoverable scientific laws. As such, an axiom is equivalent to infinitely many data observations that penalize scientific laws outside a subspace but provide no information that discriminates between scientific laws within the subspace. 

\color{black}

\subsection{Deriving the Hagen-Poiseuille Equation}\label{ssec:Hagen}
We consider the problem of deriving the velocity of laminar fluid flow through a circular pipe, from a simplified version of the Navier-Stokes equations, an assumption that the velocity can be modeled by a degree-two polynomial in the radius of the pipe, and a no-slip boundary condition. Letting $u(r)$ denote the velocity in the pipe  where $r$ is the distance from the center of the pipe, $R$ denote the radius of the pipe, $\Delta p$ denote the pressure differential throughout the pipe, $L$ denote the length of the pipe, and $\mu$ denote the viscosity of the fluid, we have the following velocity profile for $r\in [0,R]$:
\begin{align}
    u(r)=\frac{-\Delta p}{4 L \mu}(r^2-R^2).
\end{align}
We now derive this law axiomatically, by assuming that the velocity profile can be modeled by a symmetric polynomial, and iteratively increasing the degree of the polynomial until we obtain a polynomial solution, consistent with Occam's Razor. Accordingly, we initially set the degree of $u$ to be two and add together the following terms with appropriate polynomial multipliers:
\begin{align}
u=c_0+c_2 r^2, \label{eqn:functionalform}\\
\mu \frac{\partial}{\partial r}(r \frac{\partial}{\partial r} u)-r \frac{dp}{dx}=0, \label{eqn:NS_simplified}\\
c_0+c_2 R^2=0,\label{eqn:noslip}\\
L \frac{dp}{dx}=-\Delta p.\label{eqn:pressuregradient}
\end{align}
Here Equation \eqref{eqn:functionalform} posits a quadratic velocity profile in $r$, Equation \eqref{eqn:NS_simplified} imposes a simplified version of the Navier-Stokes equations in spherical coordinates, Equation \eqref{eqn:noslip} imposes a no-slip boundary condition on the velocity profile of the form $u(R)=0$, and Equation \eqref{eqn:pressuregradient} posits that the pressure gradient throughout the pipe is constant. The variables in this axiom system are $u, r, R, L, \mu, \Delta p, c_0, c_2,$ and $\frac{dp}{dx}$. We treat $c_0, c_2, \frac{dp}{dx}$ as variables that cannot be measured, and use the \verb|differentiate| function in \verb|Julia| to symbolically differentiate $u=c_0+c_2 r^2$ with respect to $r$ in Equation \eqref{eqn:NS_simplified} before solving the problem, giving the equivalent expression $4 c_2 \mu r - r \frac{dp}{dx}$. Solving Problem \eqref{prob:existence} with $u$ as the dependent variable, and searching for polynomial multipliers (and polynomial $q$) of degree at most $3$ in each variable and an overall degree of at most $6$, we  get the expression:
\begin{align*}
    4 r L \mu u - r \Delta p (R^2-r^2) = 0,
\end{align*}
which confirms the result. The associated polynomial multipliers for Equations \eqref{eqn:functionalform}--\eqref{eqn:pressuregradient} are:
\begin{align*}
    4 r L \mu,\\
    r^2 L-L R^2,\\
    4 r L \mu,\\
    r^3-r R^2.
\end{align*}

\subsection{Radiated Gravitational Wave Power}\label{ssec:gravwaves}
We now consider the problem of deriving the power radiated from gravitational waves emitted by two{\color{black}-}point masses orbiting their common center of gravity in a Keplerian orbit, as originally derived by Peters and Mathews~\cite{peters1963gravitational} and verified for binary star systems by Hulse and Taylor~\cite{hulse1975discovery}. Specifically, 
\cite{peters1963gravitational} showed that the average power generated by such a system is:
\begin{align*}
P=-\frac{32 G^4}{5 c^5 r^5}(m_1 m_2)^2(m_1+m_2),
\end{align*}
where $P$ is the (average) power of the wave, $G=6.6743 \times 10^{-11} m^3 kg^{-1} s^{-2}$ is the universal gravitational constant, $c$ is the speed of light, $m_1, m_2$ are the masses of the objects, and we assume that the two objects orbit a constant distance of $r$ away from each other. Note that this equation is one of the twenty so-called bonus laws considered in the work introducing AI-Feynman~\cite{udrescu2020ai}, and notably, is one of only two such laws that neither AI-Feynman nor Eureqa~\cite{dubvcakova2011eureqa} were able to derive.  We now derive this law axiomatically, by combining the following axioms with appropriate multipliers:
\begin{align}
\omega^2 r^3-G(m_1+m_2)=0, \label{eqn:kepler3}\\x
5 (m_1+m_2)^2 c^5 P+G \mathrm{Tr}\left(\frac{d^3}{dt^3}\left(m_1 m_2 r^2 \begin{pmatrix} x^2-\frac{1}{3} & x y & 0\\
    x y & y^2-\frac{1}{3} & 0 \\ 0 & 0 & -\frac{1}{3}
\end{pmatrix}\right)^2\right)=0, \label{eqn:waveequationpower}\\
x^2+y^2=1,\label{eqn:trigidentity}
\end{align}
where we make the variable substitution $x=\cos\phi, y=\sin \phi$, {\color{black}$\mathrm{Tr}$ stands for the trace function,} and we manually define the derivative of a bivariate degree-two trigonometric polynomial in $\phi=\phi_0+\omega t$ in $(x,y)$ in terms of $(x, y, \omega)$ as the following linear operator:
\begin{align*}
    \frac{d}{dt} \left(\begin{pmatrix} \sin \phi \\ \cos \phi\end{pmatrix}^\top \begin{pmatrix} a_{1,1} & a_{1,2} \\ a_{2,1} & a_{2,2} \end{pmatrix}  \begin{pmatrix} \sin \phi \\ \cos \phi\end{pmatrix}\right)= \omega \begin{pmatrix} \sin \phi \\ \cos \phi\end{pmatrix}^\top  \begin{pmatrix} a_{1,2}+a_{2,1} & a_{1,1}-a_{2,2}\\ a_{1,1}-a_{2,2} &  -a_{1,2}-a_{2,1} \end{pmatrix}\begin{pmatrix} \sin \phi \\ \cos \phi\end{pmatrix}. 
\end{align*}

Note that Equation \eqref{eqn:kepler3} is a restatement of Kepler's previously derived third law of planetary motion, Equation \eqref{eqn:waveequationpower} provides the gravitational power of a wave when the wavelength is large compared to the source dimensions, by linearizing the equations of general relativity, the third equation defines the quadruple moment tensor, and Equation \eqref{eqn:trigidentity} (which we state as $x^2+y^2=1$ within our axioms) is a standard trigonometric identity. Solving Problem \eqref{prob:existence} with $P$ as the dependent variable, and searching for a formula involving $P, G,  r, c, m_1, m_2$ with polynomial multipliers of degree at most $20$, and allowing each variable to be raised to a power for the variables $(P, x,y, \omega,G,r,c,m_1,m_2)$ of at most $(1,4, 4, 4, 3, 6, 1, 5,5)$ respectively, then yields the following equation:
\begin{align}
    \frac{1}{4}P r^5 c^5 (m_1+m_2)^2=\frac{-8}{5}G^4 m_1^2 m_2^2 (m_1+m_2)^3,
\end{align}
which verifies the result. Note that this equation is somewhat expensive to derive, owing to fact that searching over the set of degree $20$ polynomial multipliers necessitates generating a large number of linear equalities, and writing these equalities to memory is both time and memory intensive. Accordingly, we solved Problem \eqref{prob:existence} using the MIT SuperCloud environment \cite{reuther2018interactive} with $640$ GB RAM. The resulting system of linear inequalities involves $416392$ candidate monomials, and takes $14368$s to write the problem to memory and $6.58$s to be solved by \verb|Mosek|. This shows that the correctness of the universal gravitational wave equation can be confirmed via the following multipliers:
\begin{align}
\frac{-8}{5} G m_1^2 m_2^2\left(\omega^4 r^6 (x^2+y^2)^2+\omega^2 r^3 G (m_1+m_2)+G^2 (m_1+m_2)^2\right),\\
    \frac{1}{20}r^5,\\
    \frac{-8}{5}\omega^4 r^6 G^2 m_1^2 m_2^2 (m_1+m_2)(x^2+y^2+1).
\end{align}

\begin{figure}[h!]
    \centering
\includegraphics[width=0.8\textwidth]{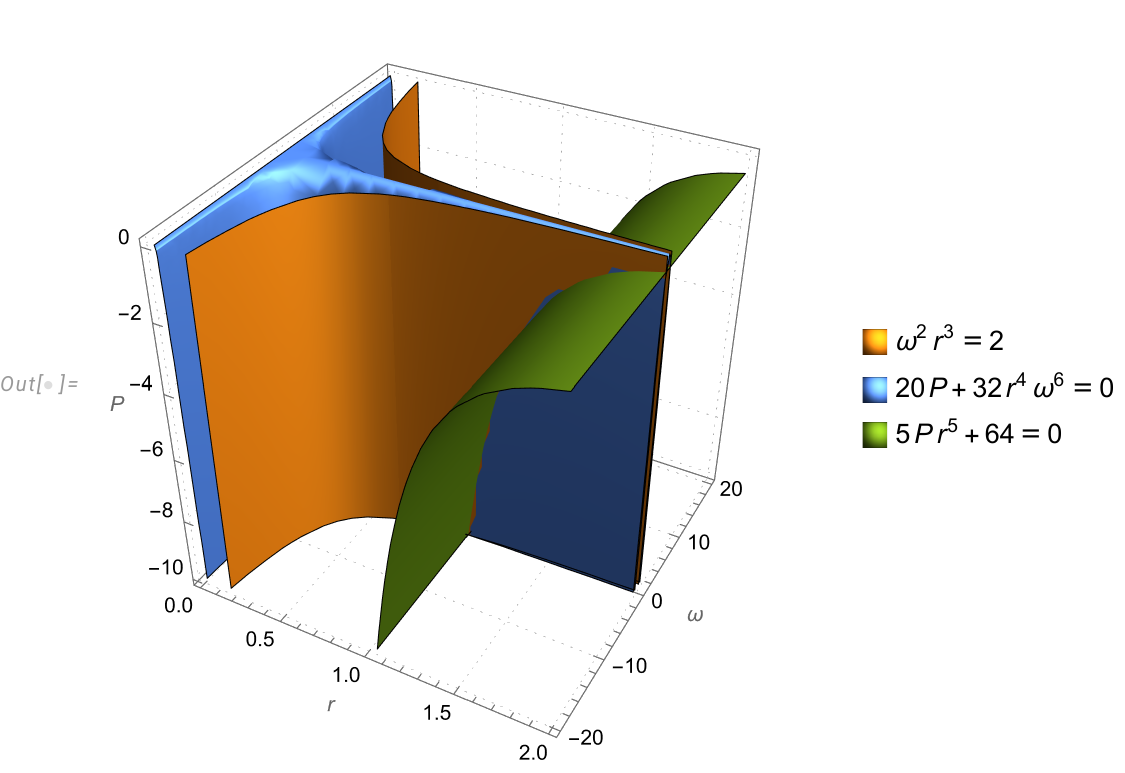}
    \caption{Illustration of the Positivstellensatz and its ability to recover the Radiation Gravitational Wave Power Equation in the special case where $m_1=m_2=c=G=1$. Keeping other variables constant, the points that obey the power equation are the intersection of the points that obey Kepler's Third Law and the points of a linearized equation from general relativity, and the wave equation is recoverable by adding these other equations with appropriate polynomial multipliers.}
    \label{fig:psatz_grav_waves}
\end{figure}

Finally, Figure~\ref{fig:psatz_grav_waves} illustrates how the Positivstellensatz derives this equation, by demonstrating that (setting $m_1=m_2=c=G=1$), the gravitational wave equation is precisely the set of points $(\omega, r, P)$ where our axioms hold with equality.

\subsection{Deriving Einstein's Relativistic Time Dilation Law}\label{ssec:einstein}

Next, we consider the problem of deriving Einstein's relativistic time dilation formula from a complete set of background knowledge axioms plus an inconsistent ``Newtonian'' axiom, which posits that light behaves like a mechanical object. We distinguish between these axioms using data on the relationship between the velocity of a light clock and the relative passage of time, as measured experimentally by Chou et at.~\cite{chou2010optical} and stated explicitly in the work of Cornelio et al.~\citep[Tab. 6]{cornelio2021ai}.

Einstein's law describes 
the relationship between how two observers in relative motion to each other observe time, and demonstrates that observers moving at different speeds experience time differently. Indeed, letting the constant $c$ denote the speed of light, the frequency $f$ of a clock moving at a speed $v$ is related to the frequency $f_0$ of a stationary clock via
\begin{align} \label{einsteins-law}
\frac{f}{f_0}=\sqrt{1-\frac{v^2}{c^2}}.
\end{align}

We now derive this law axiomatically, by adding together the following five axioms with appropriate polynomial multipliers:
\begin{align}
c dt_0-2d=0,\label{eqn:rel1}\\
c dt-2L=0, \label{eqn:rel2}\\
4 L^2+4 d^2-v^2 dt^2=0, \label{eqn:rel3}\\
f dt_0=1, \label{eqn:rel4}\\
f dt=1, \label{eqn:rel5}
\end{align}
plus the following (inconsistent) Newtonian axiom:
\begin{align}
    dt^2(v^2+c^2)-4L^2=0, \label{axiom:newtonian}
\end{align}
where $dt_0$ denotes the time required for a light to travel between two stationary mirrors separated by a distance $d$, and $dt$ denotes the time required for light to travel between two similar mirrors moving at velocity $v$, giving a distance between the mirrors of $L$.

These axioms have the following meaning: Equation \eqref{eqn:rel1} relates the time required for light to travel between two stationary mirrors to their distance, Equation \eqref{eqn:rel2} similarly relates the time required for light to travel between two mirrors in motion to the effective distance $L$, Equation \eqref{eqn:rel3} relates the physical distance between the mirrors $d$ to their effective distance $L$ induced by the motion $v$ via the Pythagorean theorem, and Equations \eqref{eqn:rel4}-\eqref{eqn:rel5} relate frequencies and periods. Finally, Equation \eqref{axiom:newtonian} assumes (incorrectly) that light behaves like other mechanical objects, meaning if it is emitted orthogonally from an object traveling at velocity $v$, then it has velocity $\sqrt{v^2+c^2}$.

By solving Problem \eqref{prob:overal} with a cardinality constraint that we include at most $\tau=5$ axioms (corresponding to the exclusion of one axiom), a constraint that we must exclude either Equation \eqref{eqn:rel2} or Equation \eqref{axiom:newtonian}, $f$ as the dependent variable, experimental data in $f, f_0, v, c$ to separate the valid and invalid axioms (obtained from \citep[Tab. 6]{cornelio2021ai} by setting $f_0=1$ to transform the data in $(f-f_0)/f_0$ into data in $f, f_0$), $f_0, v, c$ as variables that we would like to appear in our polynomial formula $q(\bm{x})=0 \ \forall \bm{x} \in \mathcal{G} \cap \mathcal{H}$, and searching the set of polynomial multipliers of degree at most $2$ in each term, we obtain the law:
\begin{align}
-c^2 f_0^2+c^2 f^2+f_0^2 v^2=0,
\end{align}
in $6.04$ seconds using \verb|Gurobi| version $9.5.1$. Moreover, we immediately recognize this as a restatement of Einstein's law \ref{einsteins-law}. This shows that the correctness of Einstein's law can be verified by multiplying the (consistent relativistic set of) axioms by the following polynomials:
\begin{align}
2 d f_0^2 f^2+c f_0 f^2,\\
-c f_0^2 f-2 f_0^2 f^2 L,\\
-f_0^2 f^2,\\
-2 c d f_0 f^2-c^2 f^2,\\
c^2 dt f_0^2 f - dt f_0^2 f v^2+c^2 f_0^2 -f_0^2 v^2.
\end{align}
Moreover, it verifies that relativistic axioms, particularly the axiom $c dt=2L$, fit the light clock data of \cite{chou2010optical} better than Newtonian axioms, because, by the definition of Problem \eqref{prob:overal}, \verb|AI-Hilbert| selects the combination of $\tau=5$ axioms with the lowest discrepancy between the discovered scientific formula and the experimental data.

\subsection{Deriving Kepler's Third Law of Planetary Motion}\label{ssec:kepler}
In this section, we consider the problem of deriving Kepler's third law of planetary motion from a complete set of background knowledge axioms plus an incorrect candidate formula {as an additional axiom}, which is to be screened out using experimental data. To our knowledge, this paper is the first work that addresses this particularly challenging problem setting. Indeed, none of the approaches to scientific discovery reviewed in the introduction successfully distinguish between correct and incorrect axioms via experimental data by solving a single optimization problem. The primary motivation for this experiment is to demonstrate that \verb|AI-Hilbert| provides a system for determining whether, given a background theory and experimental data, it is possible to improve upon a state-of-the-art scientific formula using background theory and experimental data.

Kepler's law describes the relationship between the distance between two bodies, e.g., the sun and a planet, and their orbital periods and takes the form:
\begin{align}
p=\sqrt{\frac{4\pi^2(d_1+d_2)^3}{G(m_1+m_2)}},
\end{align}
where $G=6.6743 \times 10^{-11} m^3 kg^{-1} s^{-2}$ is the universal gravitational constant,$m_1$ and $m_2$ are the masses of the two bodies, $d_1$ and $d_2$ are the respective distances between $m_1$, $m_2$ and their common center of mass, and $p$ is the orbital period. We now derive this law axiomatically by adding together the following five axioms with appropriate polynomial multipliers:
\begin{align}
d_1 m_1-d_2 m_2=0, \label{eqn:com}\\
(d_1+d_2)^2 F_g -G m_1 m_2=0, \label{eqn:grav}\\
F_c- m_2 d_2 w^2=0, \label{eqn:centri}\\
F_c-F_g=0,\label{eq:match}\\
wp=1, \label{eqn:freqperiod}
\end{align}
plus the following (incorrect) candidate formula {(as an additional axiom)} proposed by Cornelio et al.~\cite{cornelio2021ai} for the exoplanet dataset (where the mass of the planets can be discarded as negligible when added to the much bigger mass of the star):
\begin{align}
    p^2 m_1-0.1319 (d_1+d_2)^3=0~.
\end{align}
Here $F_g$ and $F_c$ denote the gravitational and centrifugal forces in the system, and $w$ denotes the frequency of revolution. Note that we replaced $p$ with $2\pi p$ in our definition of revolution period in order that $\pi$ does not feature in our equations; we divide $p$ by $2\pi$ after deriving Kepler's law.

The above axioms have the following meaning: Equation \eqref{eqn:com} defines the center of mass of the dynamical system, Equation \eqref{eqn:grav} defines the gravitational force of the system, Equation \eqref{eqn:centri} defines the centrifugal force of the system, Equation \eqref{eq:match} matches the centrifugal and dynamical forces, and Equation \eqref{eqn:freqperiod} relates the frequency and the period of revolution. 

Accordingly, we solve our polynomial optimization problem under a sparsity constraint that at most $\tau=5$ axioms can be used to derive our model, a constraint that $d^c=0$ (meaning we need not specify the hyperparameter $\lambda$ in \eqref{prob:overal}), by minimizing the objective
\begin{align*}
    \sum_{i=1}^n \vert q(\bar{\bm{x}_i})\vert,
\end{align*}
where $q$ is our implicit polynomial and $\{\bar{\bm{x}_i}\}_{i=1}^4$ is a set of observations of the revolution period of binary stars stated in \cite[Tab. 5]{cornelio2021ai}. Searching over the set of degree-five polynomials $q$ derivable using degree six certificates then yields a mixed-integer linear optimization problem in $18958$ continuous and $6$ discrete variables, with the solution:
\begin{align}\label{eqn:kepler_implicit_1}
m_1 m_2 G p^2-m_1 d_1 d_2^2-m_2 d_1^2 d_2-2 m_2 d_1 d_2^2=0,
\end{align}
which is precisely Kepler's third law. The validity of this equation can be verified by adding together our axioms with the weights:
\begin{align}
-d_2^2 p^2 w^2,\\
-p^2,\\
d_1^2 p^2 +2 d_1 d_2 p^2+d_2^2 p^2,\\
d_1^2 p^2 +2 d_1 d_2 p^2+d_2^2 p^2,\\
m_1 d_1 d_2^2 p w+m_2d_1^2 d_2 p w + 2 m_2 d_1 d_2^2 p w + m_1 d_1 d_2^2  + m_2 d_1^2 d_2 + 2 m_2 d_1 d_2^2,
\end{align}
as previously summarized in Figure~\ref{fig:system}. This is significant, because existing works on symbolic regression and scientific discovery \citep[][]{udrescu2020ai,guimera2020bayesian} often struggle to derive Kepler's law, even given observational data. Indeed, our approach is also more scalable than deriving Kepler's law manually; Johannes Kepler spent four years laboriously analyzing stellar data to arrive at his law \cite{russell1964kepler}.

\subsection{\color{black}Kepler Revisited With Missing Axioms}\label{ssec:kepler2.0}

 \color{black}

In this section, we revisit the problem of deriving Kepler's third law of planetary motion considered in Section \ref{ssec:kepler}, with a view to verify AI-Hilbert's ability to discover scientific laws from a combination of theory and data. Specifically, rather than providing a complete (albeit inconsistent) set of background theory as in Section \ref{ssec:kepler}, we suppress a subset of the axioms \eqref{eqn:com}--\eqref{eqn:freqperiod} and investigate the number of noiseless data points required to recover Equation \eqref{eqn:kepler_implicit_1}. 
To simplify our analysis, we set $G=1$ and generate noiseless data observations by sampling the values of the independent variables (the masses of the two bodies and the distance between them) uniformly at random in the ranges observed in real data (i.e., exoplanet dataset in AI-Descartes~\cite{cornelio2021ai}) and computing the value of the dependent variable (the revolution period) using the ground truth formula.

To exploit the fact that our data observations are noiseless, we solve the following variant of \eqref{prob:overal}: 
\begin{align*}
\min_{q \in \mathbb{R}_{n,2d}} \quad & \frac{\lambda_1}{\sqrt{\vert \mathcal{D}\vert}}\sum_{\bar{\bm{x}_i} \in \mathcal{D}} \vert q(\bar{\bm{x}_i})\vert+\lambda_2 \cdot d^c(q, \mathcal{G} \cap \mathcal{H})+(1-\lambda_1-\lambda_2)\Vert q \Vert_1
\\
\text{s.t.}
\quad & \sum_{\bm{\mu} \in \Omega: \bm{\mu}_1 \geq 1} a_{\bm{\mu}}=1, \nonumber\\
& a_{\bm{\mu}}=0  \ \ \forall \bm{\mu} \in \Omega: \sum_{j=t+1}^n \bm{\mu}_j \geq 1,\nonumber \\
& \sum_{\bar{\bm{x}_i} \in \mathcal{D}} \vert q(\bar{\bm{x}_i})\vert \leq \vert \mathcal{D}\vert \epsilon,\\
& \sum_{\bm{\mu} \in \Omega: \bm{\mu}_1 = 0} a_{\bm{\mu}} \leq -1/10 \ {\bigvee} \ \sum_{\bm{\mu} \in \Omega: \bm{\mu}_1 = 0} a_{\bm{\mu}} \geq 1/10 \nonumber
\end{align*}
where we set $\lambda_1=0.9, \lambda_2=0.01$, $\epsilon=10^{-7}$ for all experiments, seek a degree $4$ polynomial $q$ using a proof certificate of degree at most $6$, and use $L_1$-coefficient norm of $q$ as a regularization term  analogously to Lasso regression \cite{tibshirani1996regression}. Note that the second-to-last constraint ensures that the derived polynomial $q$ explains all (noiseless) observations up to a small tolerance. Further, the last constraint is imposed as a linear inequality constraint with an auxiliary binary variable 
via the big-$M$ technique \cite{glover1975improved, bertsimas2021unified}, to ensure that the derived formula includes at least one term not involving the rotation period.

\begin{figure}[ht]
 \centering
\begin{subfigure}[ht]{0.45\textwidth}
    \centering
    \includegraphics[width=\textwidth]{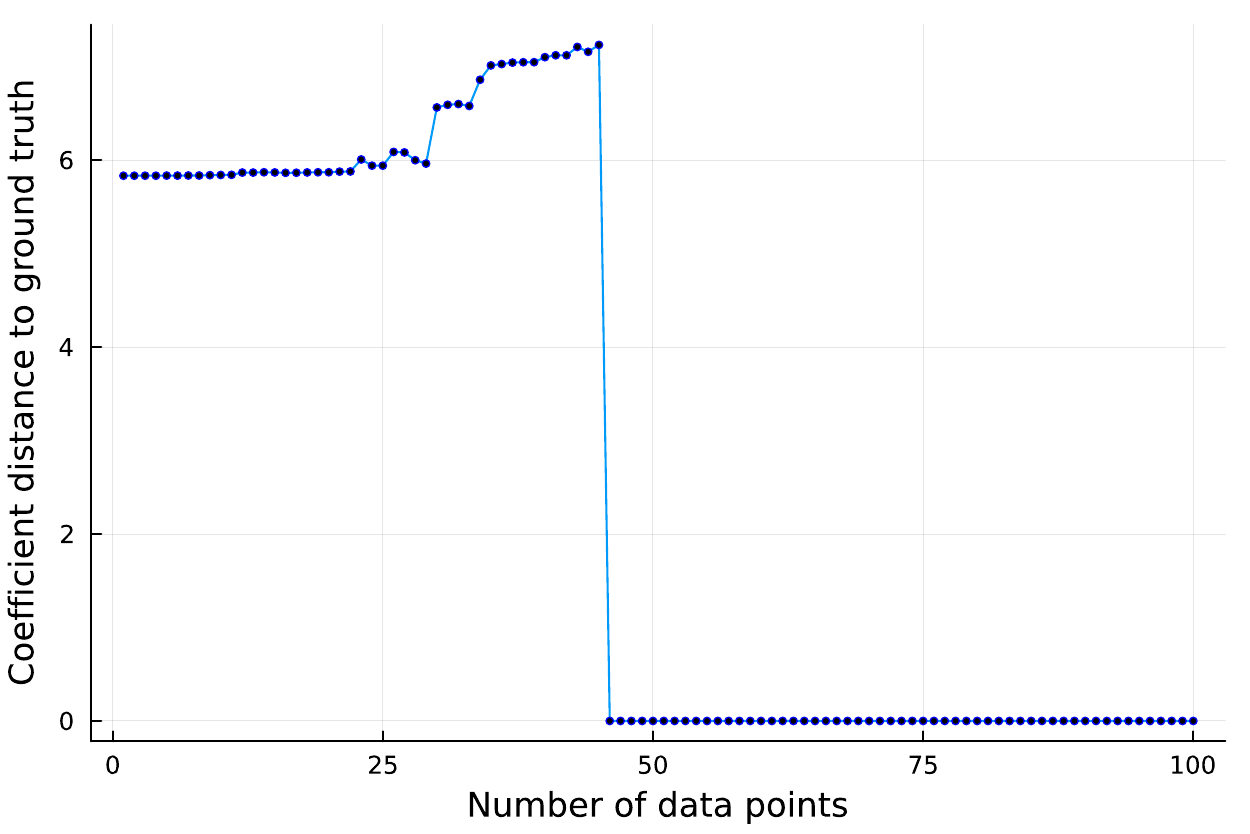}
    \subcaption{\color{black}Omitting axiom $d_1 m_1 - d_2 m_2 = 0$}
\end{subfigure}
\begin{subfigure}[ht]{0.45\textwidth}
 \centering
\includegraphics[width=\textwidth]{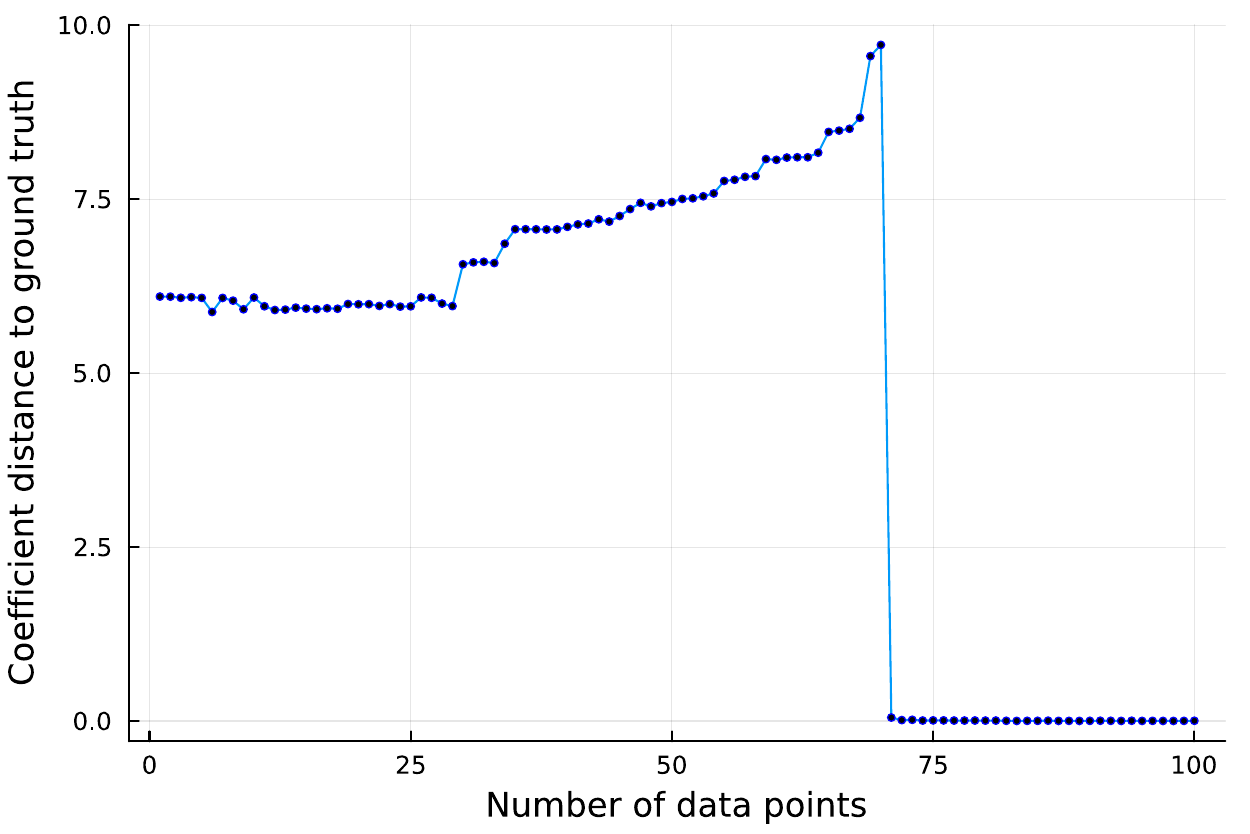}
\subcaption{\color{black}Omitting all axioms}
\end{subfigure}
\caption{\color{black}Coefficient distance between scientific formula derived by {AI-Hilbert} and ground truth vs. the number of data points when we omit some axioms (left) or all axioms (right).} 
\label{fig:kepler_data}
\end{figure}

Figure \ref{fig:kepler_data} depicts the $\ell_2$ coefficient distance between the scientific formula derived by \verb|AI-Hilbert| and Equation \eqref{eqn:kepler_implicit_1} mod $d_1 m_1-d_2 m_2=0$ as we increase the number of data points, where we suppress the axiom $d_1 m_1-d_2 m_2=0$ (left), where we suppress all axioms (right). In both cases, there is an all-or-nothing phase transition \citep[see also][for a discussion of this phenomenon throughout machine learning ]{gamarnik2021overlap} in AI-Hilbert's ability to recover the scientific law, where before a threshold number of data points, \verb|AI-Hilbert| cannot recover Kepler's law, and beyond the threshold
, \verb|AI-Hilbert| recovers the law exactly. 

Note that the increase in the coefficient distance before $m=71$ data points reflects that solutions near $q=0$ (and thus closer in coefficient norm to the ground truth) are optimal with $m=0$ data points but do not fit a small number of data points perfectly, while polynomials $q$ further from the ground truth in coefficient norm fit a small number of data points perfectly. Indeed, the total error with respect to the training data is less than $10^{-5}$ for all values of $m$ in both problem settings.

Figure \ref{fig:kepler_data} reveals that when only the axiom $d_1 m_1-d_2 m_2=0$ is missing, it is possible to perform scientific discovery with as few as $46$ data points, while at least $71$ data points are needed when all axioms are missing. This is because the axiom $d_1 m_1-d_2 m_2=0$ multiplied by the term in the proof certificate $-d_2^2 p^2 w^2$ (Section \ref{ssec:kepler}) is of a similar complexity as Kepler's Third Law. Thus, the value of $d_1 m_1-d_2 m_2=0$ is $46$ data points, while the value of all axioms is $71$ data points. The value of data compared to background theory depends, in general, on the problem setting and the data quality, as well as how well dispersed the data samples are.

\color{black}

\subsection{Bell Inequalities}\label{ssec:bell}
We now consider the problem of deriving Bell Inequalities in quantum mechanics. Bell Inequalities~\citep[]{bell1964einstein} are well-known in physics because they provide bounds on the correlation of measurements in any multi-particle system which obeys local realism (i.e., for which a joint probability distribution exists), that are violated experimentally, thus demonstrating that the natural world does not obey local realism. For ease of exposition, we prove a version called the GHZ inequality~\cite{greenberger1990bell}. Namely, given random variables $A, B, C$ which take values on $\{\pm 1\}$, for any joint probability distribution describing $A, B, C$, it follows that
\begin{align}
    \mathbb{P}(A=B)+\mathbb{P}(A=C)+\mathbb{P}(B=C) \geq 1,
\end{align}
but this bound is violated experimentally~\cite{fahmi2002locality}. 

We derive this result axiomatically, using Kolmogorov's probability axioms and the specialization of our sum-of-squares framework to linear optimization proposed in Section \ref{ssec:specialcases}, which is a valid specialization because the entire problem is linear. In particular, letting $p_{-1,1,-1}=\mathbb{P}(A=-1, B=1, C=-1)$, deriving the largest lower bound for which this inequality holds is equivalent to solving the following linear optimization problem:
\begin{equation*}
    \min ~ p_{AB}+p_{BC}+p_{AC} ~~ \text{s.t.} ~~ p \in \mathcal{S},
\end{equation*}
where $\mathcal{S}:=\{\bm{p} \geq \bm{0}, \bm{e}^\top \bm{p}=1\}$, $p_{AB}:=p_{-1,-1,-1}+p_{-1,-1,1}+p_{1,1,-1}+p_{1,1,1}$ and $p_{AC}$, $p_{BC}$ are defined similarly. 

We solve this problem using \verb|Gurobi| and \verb|Julia|, which verifies 
that $\gamma=1$ is the largest value for which this inequality holds, and obtains the desired inequality. Moreover, the solution to its dual problem yields the certificate
$2 p_{-1,-1,-1}+2 p_{1,1,1} \geq 0$
which verifies that $1$ is indeed a valid lower bound for $p_{AB}+p_{BC}+p_{AC}$, by adding $\bm{e}^\top \bm{p}$ to the left-hand side of this certificate and $1$ to the right-hand side. 

To further demonstrate the generality and utility of our approach, we now derive a more challenging Bell inequality, namely the so-called I3322 inequality (c.f.~\cite{froissart1981constructive}). Given particles $A_1, A_2, A_3, B_1, B_2, B_3$ which take values on $\{\pm 1\}$, the inequality reveals that for any joint probability distribution, we have:
\begin{align*}
    \mathbb{E}[A_1]-\mathbb{E}[A_2]+\mathbb{E}[B_1]-\mathbb{E}[B_2]&-\mathbb{E}[(A_1-A_2)(B_1-B_2)]\\
    & +\mathbb{E}[(A_1+A_2)B_3]+\mathbb{E}[A_3(B_1+B_2)] \leq 4.
\end{align*}

Using the same approach as previously, and defining $\bm{p}$ to be an arbitrary discrete probability measure on $\{\pm 1\}^6$, we verify that the smallest such upper bound which holds for each joint probability measure is $4$, with the following polynomial certificate modulo $\bm{e}^\top \bm{p}=1$:
\begin{align*}
& 4 p_{2,1,1,1,1,1}+4 p_{1,2,1,1,1,1}+8 p_{2,2,1,1,1,1}+4 p_{2,1,2,1,1,1}+4p_{1,2,2,1,1,1}+8p_{2,2,2,1,1,1}\\
& +4p_{1,1,1,2,1,1} +8p_{2,1,1,2,1,1}+4p_{1,2,1,2,1,1}+8p_{2,2,1,2,1,1}+4p_{2,1,2,2,1,1}+4p_{2,2,2,2,1,1}\\
& +4 p_{1,1,1,1,2,1}+4 p_{2,1,1,1,2,1}+12 p_{1,2,1,1,2,1}+12 p_{2,2,1,1,2,1}+8 p_{1,2,2,1,2,1}
+8p_{2,2,2,1,2,1}\\
& +8 p_{1,1,1,2,2,1}+8 p_{2,1,1,2,2,1}+12 p_{1,2,1,2,2,1} +12 p_{2,2,1,2,2,1}+4 p_{1,2,2,2,2,1}+4p_{2,2,2,2,2,1}\\
& +4p_{1,1,2,1,1,2}+4p_{2,1,2,1,1,2}+4 p_{1,2,2,1,1,2} +4p_{2,2,2,1,1,2} +4p_{1,1,1,2,1,2}+4p_{2,1,1,2,1,2}\\
& +4p_{1,1,2,2,1,2}
+4p_{2,1,2,2,1,2}+4p_{1,1,1,1,2,2} +8p_{1,2,1,1,2,2} +4p_{2,2,1,1,2,2} +4p_{1,1,2,1,2,2}\\
& +8p_{1,2,2,1,2,2}+4p_{2,2,2,1,2,2}
+8p_{1,1,1,2,2,2} + 4p_{2,1,1,2,2,2} + 8p_{1,2,1,2,2,2}+4p_{2,2,1,2,2,2}\\
& +4p_{1,1,2,2,2,2}+4p_{1,2,2,2,2,2} \geq 0
\end{align*}
where an index of $1$ denotes that a random variable took the value $-1$ and an index of $2$ denotes that a random variable took the value $1$, and the random variables are indexed in the order $A_1, A_2, A_3, B_1, B_2, B_3$.

\section{Discussion and Future Developments} \label{sec:summary} 
In this work, we proposed a new approach to scientific discovery that leverages ideas from real algebraic geometry and mixed-integer optimization to discover new scientific laws from a possibly inconsistent or incomplete set of scientific axioms and noisy experimental data. This improves existing approaches to scientific discovery that typically propose plausible scientific laws from either background theory alone or data alone. Indeed, by combining data and background theory in the discovery process, we potentially allow scientific discoveries to be made in previously inhospitable regimes where there is limited data and/or background theory, and gathering data is expensive. We hope our approach serves as an exciting tool that assists the scientific community in efficiently and accurately explaining the natural world. 

Inspired by the success of \verb|AI-Hilbert| in rediscovering existing scientific laws, we conclude by discussing some exciting research directions that are natural extensions of this work.  

\paragraph{Improving the Generality of AI-Hilbert:}
This work proposes a symbolic discovery framework that combines background theory expressible as a system of polynomial equalities and inequalities, or that can be reformulated as such a system (e.g., in a Polar coordinate system, by substituting $x=r \cos 
\theta, y=r \sin \theta$ and requiring that $x^2+y^2=r^2$). However, many scientific discovery contexts involve background theory that cannot easily be expressed via polynomial equalities and inequalities, including differential operators, integrals, and limits, among other operators. Therefore, extending \verb|AI-Hilbert| to encompass these non-polynomial settings would be of interest. 

We point out that several authors have already proposed extensions of the sum-of-squares paradigm beyond polynomial basis functions, and these works offer a promising starting point for performing such an extension. Namely, L{\"o}fberg and Parrilo \cite{lofberg2004coefficients} (see also Bach \cite{bach2022sum} and Bach and Rudi \cite{bach2023exponential}) propose an extension to trigonometric basis functions, and Fawzi et al.~\cite{fawzi2019semidefinite} propose approximating univariate non-polynomial functions via their Gaussian quadrature and Pad{\'e} approximants. Moreover, Huchette and Vielma~\cite{huchette2022nonconvex} advocate modeling non-convex functions via piecewise linear approximations with strong dual bounds. Using such polynomial approximations of non-polynomial operators offers one promising path for extending \verb|AI-Hilbert| to the non-polynomial setting. 

\paragraph{Automatically Parameter-Tuning AI-Hilbert: } \verb|AI-Hilbert| requires hyperparameter optimization by the user to trade-off the importance of fidelity to a model, fidelity to experimental data, and complexity of the symbolic model. Therefore, one extension of this work could be to automate this hyperparameter optimization process, by automatically solving mixed-integer and semidefinite optimization problems with different bounds on the degree of the proof certificates and different weights on the relative importance of fidelity to a model and fidelity to data, and using machine learning techniques to select solutions most likely to satisfy a scientist using \verb|AI-Hilbert|{\color{black}; see also \cite{yu2020hyper} for a review of automated hyperparameter optimization}.

\paragraph{Improving the Scalability of AI-Hilbert:}
One limitation of our implementation of \verb|AI-Hilbert| is that it relies on reformulating sum-of-squares optimization problems as semidefinite problems and solving them via primal-dual interior point methods (IPMs) \cite{nesterov1994interior, nesterov1997self}. This arguably presents a limitation, because the Newton step in IPMs \citep[see, e.g.,][]{alizadeh1998primal} requires performing a memory-intensive matrix inversion operation. Indeed, this matrix inversion operation is sufficiently expensive that, in our experience, \verb|AI-Hilbert| was unable to perform scientific discovery tasks with more than $n=15$ variables and a constraint on the degree of the certificates searched over of $d=20$ or greater (in general, runtime and memory usage is a function of both the number of symbolic variables and the degree of the proof certificates searched over).

To address this limitation and enhance the scalability of \verb|AI-Hilbert|, there are at least three future directions to explore. First, one could exploit ideas related to the Newton polytope (or convex hull of the exponent vectors of a polynomial) \cite{reznick1978extremal} to reduce the number of monomials in the sum-of-squares decompositions developed in this paper, as discussed in detail in \cite[Chap 3.3.4]{blekherman2012semidefinite}. Second, one could use presolving techniques such as chordal sparsity \cite{griewank1984existence, vandenberghe2015chordal} or partial facial reduction \cite{permenter2018partial, zhu2019sieve} to reduce the number of variables in the semidefinite optimization problems that arise from sum-of-squares optimization problems. Third, one could attempt to solve sum-of-squares problems without using computationally expensive interior point methods for semidefinite programs, e.g., by using a Burer-Monteiro factorization approach \cite{burer2003nonlinear, legat2022low} or by optimizing over a second-order cone inner approximation of the positive semidefinite cone \cite{ahmadi2019dsos}.

\subsection*{Acknowledgements} We are grateful to Ken Clarkson, Joao Goncalves, Phokion Kolatis, Jon Lenchner, and {\color{black}Nimrod Megiddo} (IBM) for valuable discussions on scientific discovery {\color{black}and two anonymous referees and the senior editor for valuable comments that improved the manuscript}. {\color{black}Ryan Cory-Wright and Bachir El Khadir both gratefully acknowledge IBM Research for hosting them as IBM Goldstine postdoctoral fellows while most of this work was conducted.}

\bibliographystyle{abbrv}

\begin{thebibliography}{100}

\bibitem{achterberg2019s}
T.~Achterberg.
\newblock What’s new in {G}urobi 9.0, 2019.

\bibitem{ahmadi2019polynomial}
A.~A. Ahmadi, E.~De~Klerk, and G.~Hall.
\newblock Polynomial norms.
\newblock {\em SIAM Journal on Optimization}, 29(1):399--422, 2019.

\bibitem{bachir2023sideinfo}
A.~A. Ahmadi and B.~E. Khadir.
\newblock Learning dynamical systems with side information.
\newblock {\em SIAM Review}, 65(1):183--223, 2023.

\bibitem{ahmadi2019dsos}
A.~A. Ahmadi and A.~Majumdar.
\newblock {DSOS} and {SDSOS} optimization: More tractable alternatives to sum
    of squares and semidefinite optimization.
\newblock {\em SIAM Journal on Applied Algebra and Geometry}, 3(2):193--230,
    2019.

\bibitem{alizadeh1998primal}
F.~Alizadeh, J.-P.~A. Haeberly, and M.~L. Overton.
\newblock Primal-dual interior-point methods for semidefinite programming:
    Convergence rates, stability and numerical results.
\newblock {\em SIAM Journal on Optimization}, 8(3):746--768, 1998.

\bibitem{andersen2000mosek}
E.~D. Andersen and K.~D. Andersen.
\newblock The {MOSEK} interior point optimizer for linear programming: an
    implementation of the homogeneous algorithm.
\newblock {\em High Performance Optimization}, pages 197--232, 2000.

\bibitem{arora2018decline}
A.~Arora, S.~Belenzon, and A.~Patacconi.
\newblock The decline of science in corporate {R}\&{D}.
\newblock {\em Strategic Management Journal}, 39(1):3--32, 2018.

\bibitem{arous2020free}
G.~B. Arous, A.~S. Wein, and I.~Zadik.
\newblock Free energy wells and overlap gap property in sparse {PCA}.
\newblock In {\em Conference on Learning Theory}, pages 479--482. PMLR, 2020.

\bibitem{artin1927zerlegung}
E.~Artin.
\newblock {\"U}ber die zerlegung definiter funktionen in quadrate.
\newblock In {\em Abhandlungen aus dem mathematischen Seminar der
    Universit{\"a}t Hamburg}, volume~5, pages 100--115. Springer, 1927.

\bibitem{austel2017globally}
V.~Austel, S.~Dash, O.~Gunluk, L.~Horesh, L.~Liberti, G.~Nannicini, and
    B.~Schieber.
\newblock Globally optimal symbolic regression.
\newblock {\em arXiv preprint arXiv:1710.10720}, 2017.

\bibitem{bach2022sum}
F.~Bach.
\newblock Sum-of-squares relaxations for information theory and variational
    inference.
\newblock {\em arXiv preprint arXiv:2206.13285}, 2022.

\bibitem{bach2023exponential}
F.~Bach and A.~Rudi.
\newblock Exponential convergence of sum-of-squares hierarchies for
    trigonometric polynomials.
\newblock {\em SIAM Journal on Optimization}, 33(3):2137--2159, 2023.

\bibitem{baum2021artificial}
Z.~J. Baum, X.~Yu, P.~Y. Ayala, Y.~Zhao, S.~P. Watkins, and Q.~Zhou.
\newblock Artificial intelligence in chemistry: current trends and future
    directions.
\newblock {\em Journal of Chemical Information and Modeling}, 61(7):3197--3212,
    2021.

\bibitem{bell1964einstein}
J.~S. Bell.
\newblock On the {E}instein {P}odolsky {R}osen paradox.
\newblock {\em Physics Physique Fizika}, 1(3):195, 1964.

\bibitem{bertsimas2023matrix}
D.~Bertsimas, R.~Cory-Wright, S.~Lo, and J.~Pauphilet.
\newblock Optimal low-rank matrix completion: Semidefinite relaxations and
    eigenvector disjunctions.
\newblock {\em arXiv preprint arXiv:2305.12292}, 2023.

\bibitem{bertsimas2021unified}
D.~Bertsimas, R.~Cory-Wright, and J.~Pauphilet.
\newblock A unified approach to mixed-integer optimization problems with
    logical constraints.
\newblock {\em SIAM Journal on Optimization}, 31(3):2340--2367, 2021.

\bibitem{bertsimas2022mixed}
D.~Bertsimas, R.~Cory-Wright, and J.~Pauphilet.
\newblock Mixed-projection conic optimization: A new paradigm for modeling rank
    constraints.
\newblock {\em Operations Research}, 70(6):3321--3344, 2022.

\bibitem{bertsimas2019machine}
D.~Bertsimas and J.~Dunn.
\newblock {\em Machine learning under a modern optimization lens}.
\newblock Dynamic Ideas Press, 2019.

\bibitem{bertsimas2023learning}
D.~Bertsimas and W.~Gurnee.
\newblock Learning sparse nonlinear dynamics via mixed-integer optimization.
\newblock {\em Nonlinear Dynamics}, pages 1--20, 2023.

\bibitem{bertsimas2016best}
D.~Bertsimas, A.~King, and R.~Mazumder.
\newblock {Best subset selection via a modern optimization lens}.
\newblock {\em The Annals of Statistics}, 44(2):813 -- 852, 2016.

\bibitem{bhattacharya2020stagnation}
J.~Bhattacharya and M.~Packalen.
\newblock Stagnation and scientific incentives.
\newblock Technical report, National Bureau of Economic Research, 2020.

\bibitem{bixby2007progress}
R.~Bixby and E.~Rothberg.
\newblock Progress in computational mixed integer programming--a look back from
    the other side of the tipping point.
\newblock {\em Annals of Operations Research}, 149(1):37, 2007.

\bibitem{blekherman2012semidefinite}
G.~Blekherman, P.~A. Parrilo, and R.~R. Thomas.
\newblock {\em Semidefinite optimization and convex algebraic geometry}.
\newblock SIAM, 2012.

\bibitem{bloom2020ideas}
N.~Bloom, C.~I. Jones, J.~Van~Reenen, and M.~Webb.
\newblock Are ideas getting harder to find?
\newblock {\em American Economic Review}, 110(4):1104--1144, 2020.

\bibitem{bongard2007automated}
J.~Bongard and H.~Lipson.
\newblock Automated reverse engineering of nonlinear dynamical systems.
\newblock {\em Proceedings of the National Academy of Sciences},
    104(24):9943--9948, 2007.

\bibitem{brunton2016discovering}
S.~L. Brunton, J.~L. Proctor, and J.~N. Kutz.
\newblock Discovering governing equations from data by sparse identification of
    nonlinear dynamical systems.
\newblock {\em Proceedings of the National Academy of Sciences},
    113(15):3932--3937, 2016.

\bibitem{brynjolfsson2018artificial}
E.~Brynjolfsson, D.~Rock, and C.~Syverson.
\newblock Artificial intelligence and the modern productivity paradox: A clash
    of expectations and statistics.
\newblock In {\em The economics of artificial intelligence: An agenda}, pages
    23--57. University of Chicago Press, 2018.

\bibitem{burer2003nonlinear}
S.~Burer and R.~D. Monteiro.
\newblock A nonlinear programming algorithm for solving semidefinite programs
    via low-rank factorization.
\newblock {\em Mathematical Programming}, 95(2):329--357, 2003.

\bibitem{candes2010matrix}
E.~J. Candes and Y.~Plan.
\newblock Matrix completion with noise.
\newblock {\em Proceedings of the IEEE}, 98(6):925--936, 2010.

\bibitem{chou2010optical}
C.-W. Chou, D.~B. Hume, T.~Rosenband, and D.~J. Wineland.
\newblock Optical clocks and relativity.
\newblock {\em Science}, 329(5999):1630--1633, 2010.

\bibitem{clegg1996using}
M.~Clegg, J.~Edmonds, and R.~Impagliazzo.
\newblock Using the {G}r{\"o}ebner basis algorithm to find proofs of
    unsatisfiability.
\newblock In {\em Proceedings of the twenty-eighth annual ACM symposium on
    Theory of computing}, pages 174--183, 1996.

\bibitem{cornelio2021ai}
C.~Cornelio, S.~Dash, V.~Austel, T.~Josephson, J.~Goncalves, K.~Clarkson,
    N.~Megiddo, B.~E. Khadir, and L.~Horesh.
\newblock Combining data and theory for derivable scientific discovery with
    {AI}-{D}escartes.
\newblock {\em Nature Communications}, 14(1777), 2023.

\bibitem{cowen2011great}
T.~Cowen.
\newblock {\em The great stagnation: How America ate all the low-hanging fruit
    of modern history, got sick, and will (eventually) feel better: A Penguin
    eSpecial from Dutton}.
\newblock Penguin, 2011.

\bibitem{cox2013ideals}
D.~Cox, J.~Little, and D.~O'Shea.
\newblock {\em Ideals, varieties, and algorithms: An introduction to
    computational algebraic geometry and commutative algebra}.
\newblock Springer Science \& Business Media, 2013.

\bibitem{cozad2018global}
A.~Cozad and N.~V. Sahinidis.
\newblock A global {MINLP} approach to symbolic regression.
\newblock {\em Mathematical Programming}, 170:97--119, 2018.

\bibitem{cranmer10pysr}
M.~Cranmer.
\newblock Py{SR}: Fast \& parallelized symbolic regression in {Python}/{Julia}.
\newblock \url{doi.org/10.5281/zenodo.4041459}, Sept. 2020.

\bibitem{cranmer2020discovering}
M.~Cranmer, A.~Sanchez-Gonzalez, P.~Battaglia, R.~Xu, K.~Cranmer, D.~Spergel,
    and S.~Ho.
\newblock Discovering symbolic models from deep learning with inductive biases.
\newblock {\em NeurIPS 2020}, 2020.

\bibitem{curmei2020shape}
M.~Curmei and G.~Hall.
\newblock Shape-constrained regression using sum of squares polynomials.
\newblock {\em Operations Research}, 2023.

\bibitem{de2020understanding}
H.~W. De~Regt.
\newblock Understanding, values, and the aims of science.
\newblock {\em Philosophy of Science}, 87(5):921--932, 2020.

\bibitem{dey2021branch}
S.~S. Dey, Y.~Dubey, and M.~Molinaro.
\newblock Branch-and-bound solves random binary ips in polytime.
\newblock In {\em Proceedings of the 2021 ACM-SIAM Symposium on Discrete
    Algorithms (SODA)}, pages 579--591. SIAM, 2021.

\bibitem{dirac1978directions}
P.~A. Dirac.
\newblock Directions in physics. {L}ectures delivered during a visit to
    {A}ustralia and {N}ew {Z}ealand, {A}ugust/{S}eptember 1975.
\newblock 1978.

\bibitem{dubvcakova2011eureqa}
R.~Dub{\v{c}}{\'a}kov{\'a}.
\newblock Eureqa: software review, 2011.

\bibitem{engle2022deterministic}
M.~R. Engle and N.~V. Sahinidis.
\newblock Deterministic symbolic regression with derivative information:
    General methodology and application to equations of state.
\newblock {\em AIChE Journal}, 68(6):e17457, 2022.

\bibitem{fahmi2002locality}
A.~Fahmi.
\newblock Locality, {B}ell's inequality and the {GHZ} theorem.
\newblock {\em Physics Letters A}, 303(1):1--6, 2002.

\bibitem{fawzi2019learning}
A.~Fawzi, M.~Malinowski, H.~Fawzi, and O.~Fawzi.
\newblock Learning dynamic polynomial proofs.
\newblock {\em Advances in Neural Information Processing Systems}, 32, 2019.

\bibitem{fawzi2019semidefinite}
H.~Fawzi, J.~Saunderson, and P.~A. Parrilo.
\newblock Semidefinite approximations of the matrix logarithm.
\newblock {\em Foundations of Computational Mathematics}, 19:259--296, 2019.

\bibitem{feynman1965feynman}
R.~P. Feynman, R.~B. Leighton, and M.~Sands.
\newblock The {F}eynman lectures on physics; vol. i.
\newblock {\em American Journal of Physics}, 33(9):750--752, 1965.

\bibitem{froissart1981constructive}
M.~Froissart.
\newblock Constructive generalization of {B}ell's inequalities.
\newblock {\em Nuovo Cimento B;(Italy)}, 64(2), 1981.

\bibitem{fujinuma2022big}
N.~Fujinuma, B.~DeCost, J.~Hattrick-Simpers, and S.~E. Lofland.
\newblock Why big data and compute are not necessarily the path to big
    materials science.
\newblock {\em Communications Materials}, 3(1):59, 2022.

\bibitem{fulton2015keymaera}
N.~Fulton, S.~Mitsch, J.-D. Quesel, M.~V{\"o}lp, and A.~Platzer.
\newblock {KeYmaera X}: An axiomatic tactical theorem prover for hybrid
    systems.
\newblock In {\em Automated Deduction-CADE-25: 25th International Conference on
    Automated Deduction, Berlin, Germany, August 1-7, 2015, Proceedings 25},
    pages 527--538. Springer, 2015.

\bibitem{gamarnik2021overlap}
D.~Gamarnik.
\newblock The overlap gap property: A topological barrier to optimizing over
    random structures.
\newblock {\em Proceedings of the National Academy of Sciences},
    118(41):e2108492118, 2021.

\bibitem{gamarnik2017high}
D.~Gamarnik and I.~Zadik.
\newblock High dimensional regression with binary coefficients. estimating
    squared error and a phase transtition.
\newblock In {\em Conference on Learning Theory}, pages 948--953. PMLR, 2017.

\bibitem{glover1975improved}
F.~Glover.
\newblock Improved linear integer programming formulations of nonlinear integer
    problems.
\newblock {\em Management Science}, 22(4):455--460, 1975.

\bibitem{greenberger1990bell}
D.~M. Greenberger, M.~A. Horne, A.~Shimony, and A.~Zeilinger.
\newblock Bell’s theorem without inequalities.
\newblock {\em American Journal of Physics}, 58(12):1131--1143, 1990.

\bibitem{griewank1984existence}
A.~Griewank and P.~L. Toint.
\newblock On the existence of convex decompositions of partially separable
    functions.
\newblock {\em Mathematical Programming}, 28:25--49, 1984.

\bibitem{guimera2020bayesian}
R.~Guimer{\`a}, I.~Reichardt, A.~Aguilar-Mogas, F.~A. Massucci, M.~Miranda,
    J.~Pallar{\`e}s, and M.~Sales-Pardo.
\newblock A {B}ayesian machine scientist to aid in the solution of challenging
    scientific problems.
\newblock {\em Science Advances}, 6(5):eaav6971, 2020.

\bibitem{guntuboyina2018nonparametric}
A.~Guntuboyina and B.~Sen.
\newblock Nonparametric shape-restricted regression.
\newblock {\em Statistical Science}, 33(4):568--594, 2018.

\bibitem{gupta2022branch}
S.~D. Gupta, B.~P. Van~Parys, and E.~K. Ryu.
\newblock Branch-and-bound performance estimation programming: A unified
    methodology for constructing optimal optimization methods.
\newblock {\em Mathematical Programming}, 2023.

\bibitem{hall2019engineering}
G.~Hall.
\newblock Applications of sums of squares polynomials.
\newblock In P.~Parrilo and R.~Thomas, editors, {\em Sum of Squares: Theory and
    Applications}, volume~77. Proceedings of Symposia in Applied Mathematics,
    2020.

\bibitem{hastie2009elements}
T.~Hastie, R.~Tibshirani, J.~H. Friedman, and J.~H. Friedman.
\newblock {\em The elements of statistical learning: data mining, inference,
    and prediction}, volume~2.
\newblock Springer, 2009.

\bibitem{hilbert1888darstellung}
D.~Hilbert.
\newblock {\"U}ber die darstellung definiter formen als summe von
    formenquadraten.
\newblock {\em Mathematische Annalen}, 32(3):342--350, 1888.

\bibitem{hilbert2019mathematical}
D.~Hilbert.
\newblock Mathematical problems.
\newblock In {\em Mathematics}, pages 273--278. Chapman and Hall/CRC, 2019.

\bibitem{huchette2022nonconvex}
J.~Huchette and J.~P. Vielma.
\newblock Nonconvex piecewise linear functions: Advanced formulations and
    simple modeling tools.
\newblock {\em Operations Research}, 2022.

\bibitem{hulse1975discovery}
R.~A. Hulse and J.~H. Taylor.
\newblock Discovery of a pulsar in a binary system.
\newblock {\em The Astrophysical Journal}, 195:L51--L53, 1975.

\bibitem{iten2020discovering}
R.~Iten, T.~Metger, H.~Wilming, L.~Del~Rio, and R.~Renner.
\newblock Discovering physical concepts with neural networks.
\newblock {\em Physical Review Letters}, 124(1):010508, 2020.

\bibitem{johnstone2009consistency}
I.~M. Johnstone and A.~Y. Lu.
\newblock On consistency and sparsity for principal components analysis in high
    dimensions.
\newblock {\em Journal of the American Statistical Association},
    104(486):682--693, 2009.

\bibitem{jumper2021highly}
J.~Jumper, R.~Evans, A.~Pritzel, T.~Green, M.~Figurnov, O.~Ronneberger,
    K.~Tunyasuvunakool, R.~Bates, A.~{\v{Z}}{\'\i}dek, A.~Potapenko, et~al.
\newblock Highly accurate protein structure prediction with alphafold.
\newblock {\em Nature}, 596(7873):583--589, 2021.

\bibitem{karagiorgi2022machine}
G.~Karagiorgi, G.~Kasieczka, S.~Kravitz, B.~Nachman, and D.~Shih.
\newblock Machine learning in the search for new fundamental physics.
\newblock {\em Nature Reviews Physics}, 4(6):399--412, 2022.

\bibitem{kitano2021nobel}
H.~Kitano.
\newblock Nobel turing challenge: creating the engine for scientific discovery.
\newblock {\em npj Systems Biology and Applications}, 7(1):29, 2021.

\bibitem{krivine1964anneaux}
J.-L. Krivine.
\newblock Anneaux pr{\'e}ordonn{\'e}s.
\newblock {\em Journal d'analyse math{\'e}matique}, 12:p--307, 1964.

\bibitem{kubalik2020symbolic}
J.~Kubal{\'\i}k, E.~Derner, and R.~Babu{\v{s}}ka.
\newblock Symbolic regression driven by training data and prior knowledge.
\newblock In {\em Proceedings of the 2020 Genetic and Evolutionary Computation
    Conference}, pages 958--966, 2020.

\bibitem{kubalik2021multi}
J.~Kubal{\'\i}k, E.~Derner, and R.~Babu{\v{s}}ka.
\newblock Multi-objective symbolic regression for physics-aware dynamic
    modeling.
\newblock {\em Expert Systems with Applications}, 182:115210, 2021.

\bibitem{landajuela2022unified}
M.~Landajuela, C.~S. Lee, J.~Yang, R.~Glatt, C.~P. Santiago, I.~Aravena,
    T.~Mundhenk, G.~Mulcahy, and B.~K. Petersen.
\newblock A unified framework for deep symbolic regression.
\newblock {\em Advances in Neural Information Processing Systems},
    35:33985--33998, 2022.

\bibitem{lasserre2001global}
J.~B. Lasserre.
\newblock Global optimization with polynomials and the problem of moments.
\newblock {\em SIAM Journal on Optimization}, 11(3):796--817, 2001.

\bibitem{lasserre2007sum}
J.~B. Lasserre.
\newblock A sum of squares approximation of nonnegative polynomials.
\newblock {\em SIAM Review}, 49(4):651--669, 2007.

\bibitem{laurent2009sums}
M.~Laurent.
\newblock Sums of squares, moment matrices and optimization over polynomials.
\newblock In {\em Emerging Applications of Algebraic Geometry}, pages 157--270.
    Springer, 2009.

\bibitem{laurent2005semidefinite}
M.~Laurent and F.~Rendl.
\newblock Semidefinite programming and integer programming.
\newblock {\em Handbooks in Operations Research and Management Science},
    12:393--514, 2005.

\bibitem{legat2022low}
B.~Legat, C.~Yuan, and P.~Parrilo.
\newblock Low-rank univariate sum of squares has no spurious local minima.
\newblock {\em SIAM Journal on Optimization}, 33(3):2041--2061, 2023.

\bibitem{lim2012consistency}
E.~Lim and P.~W. Glynn.
\newblock Consistency of multidimensional convex regression.
\newblock {\em Operations Research}, 60(1):196--208, 2012.

\bibitem{liu2024okridge}
J.~Liu, S.~Rosen, C.~Zhong, and C.~Rudin.
\newblock Okridge: Scalable optimal k-sparse ridge regression.
\newblock {\em Advances in Neural Information Processing Systems}, 36, 2024.

\bibitem{lofberg2004coefficients}
J.~Lofberg and P.~A. Parrilo.
\newblock From coefficients to samples: A new approach to {SOS} optimization.
\newblock In {\em 2004 43rd IEEE Conference on Decision and Control (CDC)(IEEE
    Cat. No. 04CH37601)}, volume~3, pages 3154--3159. IEEE, 2004.

\bibitem{lubin2023jump}
M.~Lubin, O.~Dowson, J.~D. Garcia, J.~Huchette, B.~Legat, and J.~P. Vielma.
\newblock Jump 1.0: recent improvements to a modeling language for mathematical
    optimization.
\newblock {\em Mathematical Programming Computation}, pages 1--9, 2023.

\bibitem{matsubara2022srsd}
Y.~Matsubara, N.~Chiba, R.~Igarashi, and Y.~Ushiku.
\newblock {SRSD}: Rethinking datasets of symbolic regression for scientific
    discovery.
\newblock In {\em NeurIPS 2022 AI for Science: Progress and Promises}, 2022.

\bibitem{nesterov1994interior}
Y.~Nesterov and A.~Nemirovskii.
\newblock {\em Interior-point polynomial algorithms in convex programming}.
\newblock SIAM, 1994.

\bibitem{nesterov1997self}
Y.~E. Nesterov and M.~J. Todd.
\newblock Self-scaled barriers and interior-point methods for convex
    programming.
\newblock {\em Mathematics of Operations Research}, 22(1):1--42, 1997.

\bibitem{open2023gpt4}
OpenAI.
\newblock {GPT-4} technical report.
\newblock {\em arXiv preprint arXiv:2303.08774}, 2023.

\bibitem{parrilo2003semidefinite}
P.~A. Parrilo.
\newblock Semidefinite programming relaxations for semialgebraic problems.
\newblock {\em Mathematical Programming}, 96(2):293--320, 2003.

\bibitem{permenter2018partial}
F.~Permenter and P.~Parrilo.
\newblock Partial facial reduction: Simplified, equivalent {SDP}s via
    approximations of the {PSD} cone.
\newblock {\em Mathematical Programming}, 171:1--54, 2018.

\bibitem{peters1963gravitational}
P.~C. Peters and J.~Mathews.
\newblock Gravitational radiation from point masses in a {K}eplerian orbit.
\newblock {\em Physical Review}, 131(1):435, 1963.

\bibitem{putinar1993positive}
M.~Putinar.
\newblock Positive polynomials on compact semi-algebraic sets.
\newblock {\em Indiana University Mathematics Journal}, 42(3):969--984, 1993.

\bibitem{ramana1997exact}
M.~V. Ramana.
\newblock An exact duality theory for semidefinite programming and its
    complexity implications.
\newblock {\em Mathematical Programming}, 77:129--162, 1997.

\bibitem{renegar2001mathematical}
J.~Renegar.
\newblock {\em A mathematical view of interior-point methods in convex
    optimization}.
\newblock SIAM, 2001.

\bibitem{reuther2018interactive}
A.~Reuther, J.~Kepner, C.~Byun, S.~Samsi, W.~Arcand, D.~Bestor, B.~Bergeron,
    V.~Gadepally, M.~Houle, M.~Hubbell, M.~Jones, A.~Klein, L.~Milechin,
    J.~Mullen, A.~Prout, A.~Rosa, C.~Yee, and P.~Michaleas.
\newblock Interactive supercomputing on 40,000 cores for machine learning and
    data analysis.
\newblock In {\em 2018 IEEE High Performance extreme Computing Conference
    (HPEC)}, pages 1--6. IEEE, 2018.

\bibitem{reznick1978extremal}
B.~Reznick.
\newblock Extremal {PSD} forms with few terms.
\newblock {\em Duke Mathematical Journal}, 45(2):363--374, 1978.

\bibitem{rudin2019stop}
C.~Rudin.
\newblock Stop explaining black box machine learning models for high stakes
    decisions and use interpretable models instead.
\newblock {\em Nature Machine Intelligence}, 1(5):206--215, 2019.

\bibitem{rudy2017data}
S.~H. Rudy, S.~L. Brunton, J.~L. Proctor, and J.~N. Kutz.
\newblock Data-driven discovery of partial differential equations.
\newblock {\em Science Advances}, 3(4):e1602614, 2017.

\bibitem{russell1964kepler}
J.~L. Russell.
\newblock Kepler's laws of planetary motion: 1609--1666.
\newblock {\em The British Journal for the History of Science}, 2(1):1--24,
    1964.

\bibitem{schmidt2009distilling}
M.~Schmidt and H.~Lipson.
\newblock Distilling free-form natural laws from experimental data.
\newblock {\em Science}, 324(5923):81--85, 2009.

\bibitem{schmidt2009symbolic}
M.~Schmidt and H.~Lipson.
\newblock Symbolic regression of implicit equations.
\newblock In {\em Genetic Programming Theory and Practice VII}, pages 73--85.
    Springer, 2009.

\bibitem{schmudgen1991moment}
K.~Schm{\"u}dgen.
\newblock The moment problem on compact semi-algebraic sets.
\newblock {\em Mathematische Annalen}, 289:283--313, 1991.

\bibitem{simon1973does}
H.~A. Simon.
\newblock Does scientific discovery have a logic?
\newblock {\em Philosophy of Science}, 40(4):471--480, 1973.

\bibitem{skajaa2015homogeneous}
A.~Skajaa and Y.~Ye.
\newblock A homogeneous interior-point algorithm for nonsymmetric convex conic
    optimization.
\newblock {\em Mathematical Programming}, 150(2):391--422, 2015.

\bibitem{stengle1974nullstellensatz}
G.~Stengle.
\newblock A {N}ullstellensatz and a {P}ositivstellensatz in semialgebraic
    geometry.
\newblock {\em Mathematische Annalen}, 207(2):87--97, 1974.

\bibitem{tibshirani1996regression}
R.~Tibshirani.
\newblock Regression shrinkage and selection via the lasso.
\newblock {\em Journal of the Royal Statistical Society Series B: Statistical
    Methodology}, 58(1):267--288, 1996.

\bibitem{AI-Feynman2.0}
S.~Udrescu, A.~Tan, J.~Feng, O.~Neto, T.~Wu, and M.~Tegmark.
\newblock {AI} {F}eynman 2.0: Pareto-optimal symbolic regression exploiting
    graph modularity.
\newblock In H.~Larochelle, M.~Ranzato, R.~Hadsell, M.~Balcan, and H.~Lin,
    editors, {\em Advances in Neural Information Processing Systems 33: Annual
    Conference on Neural Information Processing Systems 2020, NeurIPS 2020,
    December 6-12, 2020, virtual}, 2020.

\bibitem{udrescu2020ai2}
S.-M. Udrescu, A.~Tan, J.~Feng, O.~Neto, T.~Wu, and M.~Tegmark.
\newblock {AI} {F}eynman 2.0: {P}areto-optimal symbolic regression exploiting
    graph modularity.
\newblock {\em Advances in Neural Information Processing Systems},
    33:4860--4871, 2020.

\bibitem{udrescu2020ai}
S.-M. Udrescu and M.~Tegmark.
\newblock {AI} {F}eynman: A physics-inspired method for symbolic regression.
\newblock {\em Science Advances}, 6(16), 2020.

\bibitem{AI-Feynman}
S.-M. Udrescu and M.~Tegmark.
\newblock {AI} {F}eynman: A physics-inspired method for symbolic regression.
\newblock {\em Science Advances}, 2020.

\bibitem{vandenberghe2015chordal}
L.~Vandenberghe, M.~S. Andersen, et~al.
\newblock Chordal graphs and semidefinite optimization.
\newblock {\em Foundations and Trends{\textregistered} in Optimization},
    1(4):241--433, 2015.

\bibitem{wang2023scientific}
H.~Wang, T.~Fu, Y.~Du, W.~Gao, K.~Huang, Z.~Liu, P.~Chandak, S.~Liu,
    P.~Van~Katwyk, A.~Deac, et~al.
\newblock Scientific discovery in the age of artificial intelligence.
\newblock {\em Nature}, 620(7972):47--60, 2023.

\bibitem{wang2010information}
W.~Wang, M.~J. Wainwright, and K.~Ramchandran.
\newblock Information-theoretic limits on sparse signal recovery: Dense versus
    sparse measurement matrices.
\newblock {\em IEEE Transactions on Information Theory}, 56(6):2967--2979,
    2010.

\bibitem{yu2020hyper}
T.~Yu and H.~Zhu.
\newblock Hyper-parameter optimization: A review of algorithms and
    applications.
\newblock {\em arXiv preprint arXiv:2003.05689}, 2020.

\bibitem{zhao2023hausdorff}
W.~Zhao and G.~Zhou.
\newblock Hausdorff distance between convex semialgebraic sets.
\newblock {\em Journal of Global Optimization}, pages 1--21, 2023.

\bibitem{zhu2019sieve}
Y.~Zhu, G.~Pataki, and Q.~Tran-Dinh.
\newblock Sieve-{SDP}: A simple facial reduction algorithm to preprocess
    semidefinite programs.
\newblock {\em Mathematical Programming Computation}, 11:503--586, 2019.

\end{thebibliography}

\appendix

\color{black}

\section{\color{black}Comparison with Methods from the Literature}\label{append:sota_comparison}

To illustrate our system's capabilities, Table~\ref{tab:comparison_scientificlaws} compares \verb|AI-Hilbert| with four approaches from the literature in terms of their ability to recover various scientific laws from background theory and experimental data. 

The data for the first two problems comes from real measurements (denoted with ``R'' in the table) while for the other problems we rely on simulated data (denoted with ``S'' in the table). The data is created similarly to the approach used in the literature (see \cite{cornelio2021ai} or \cite{AI-Feynman} for more details): we sample ten data points uniformly at random for each problem, independently for each variable in a given continuous range of values. The range is either defined following ranges in real measurements or set artificially (e.g., $(0,1]$). We then add 1\% Gaussian noise to the dependent variable (following~\cite{AI-Feynman}), thereby creating small, noisy datasets to resemble real-life data. 

We used the background theory provided by Cornelio et al.~\cite{cornelio2021ai} for the first eight problems and  manually extracted the axioms for the remaining ones.

The systems we consider are:
\begin{itemize}
    \item {\bf AI-Descartes}~\cite{cornelio2021ai} is a system that combines logical reasoning with symbolic regression to generate meaningful formulae explaining a given phenomenon. It starts by generating a set of candidate models from data alone using sumbolic regression and then validates and ranks them using a background theory. In this way the system ensure that the chosen hypothesis is not only close to the data but also respects the laws already known about the environment, often providing a logical proof of its correctness. However, in the experiments below, we only run the symbolic regression module of AI-Descartes.
    \item {\bf AI Feynman}~\cite{udrescu2020ai, AI-Feynman2.0} is a symbolic regression approach integrating deep learning methodologies with the exploitation of some features commonly found in physics functions including unit presence, utilization of low-order polynomials, and symmetry. The algorithm comprises several modules handling polynomial fitting, exhaustive enumeration, neural network fitting, among other tasks, ultimately generating a list of potential formulas.
    \item {\bf PySR}~\cite{cranmer10pysr,cranmer2020discovering} employs a combination of regularized evolution, simulated annealing, and gradient-free optimization techniques to search for equations that best match the provided input data.
    \item {\bf Bayesian Machine Scientist (BMS)}~\cite{guimera2020bayesian} utilizes a Markov chain Monte Carlo approach to traverse the potential model space, leveraging prior expectations learned from a large empirical corpus of mathematical expressions.   
\end{itemize}

The above systems return a score (in some cases this corresponds to the error) and the complexity for each output formula. The systems PySR and Bayesian Machine Scientist return a best candidate. For AI-Descartes and AI-Feynman, we declare success if the correct formula is among the candidates returned, whereas for PySR and Bayesian Machine Scientist, we check if the best candidate formula is correct.

\subsection{Experimental results for comparison with prior literature}
We now compare \verb|AI-Hilbert| with four methods from the literature in Table~\ref{tab:comparison_scientificlaws}. Note that all four existing methods that we compare \verb|AI-Hilbert| against require experimental data to successfully discover scientific laws; further, we take results for these methods for the FSRD problems, the Relativistic time dilation problem,  and Kepler's third law from \cite{cornelio2021ai}. On the other hand,
\verb|AI-Hilbert| combines (potentially limited and noisy) data and (potentially corrupted) theory, including the two limiting cases where either only data or only theory are supplied.

\verb|AI-Hilbert| outperforms the 
existing systems, accurately recovering formulas in all studied problems.
Moreover it achieves this with less data, 
by leveraging 
relevant background theory.
Finally, our method is the only one capable of recovering symbolic constants (while the other systems combine them into single or multiple real values) thereby preserving the meaning and semantics of each individual constant.

We remark that for some methods from the literature, inexact recovery (denoted with a $\checkmark^\star$) may occur because these approaches are data-driven, and thus usually cannot derive symbolic constants associated with scientific laws. Instead, they group information concerning multiple constants into one that is less interpretable. For instance, consider the Compton Scattering problem (\ref{ssec:compton}), the correct formula is
  $l_2  = l_1 + \frac{h}{m_e  c }(1 - cos(\theta))$
  while data-driven methods usually only derive 
  $l_2  = l_1 + k(1- cos(\theta))$ (where $k$ is a real number).
  Moreover, sometimes integers are reported with a small numerical error (e.g., $1,001$ or $0.999$ instead of $1$).

\begin{table}[h]
  \caption{\color{black}Comparison of {AI-Hilbert} with state-of-the-art discovery methods in terms of ability to recover scientific laws from background theory and experimental data.
  }
  \centering\footnotesize
  \resizebox{\linewidth}{!}{
  \begin{tabular}{c l c c c c c c}
    \toprule
     Data & & {AI-Hilbert}  & AI-Descartes \cite{cornelio2021ai} & AI-Feynman \cite{udrescu2020ai}  & PySR \cite{cranmer10pysr} & BMS \cite{guimera2020bayesian}\\
    \midrule
    S & \textcolor{blue}{[\ref{ssec:Hagen}] Hagen Poiseuille}  & \textcolor{teal}{$\text{\checkmark}$}  &  \textcolor{red}{$\text{\xmark}\phantom{0}$}  & \textcolor{red}{$\text{\xmark}$} & \textcolor{red}{$\text{\xmark}$} &  \textcolor{red}{$\text{\xmark}$} \\
    S & \textcolor{blue}{[\ref{ssec:gravwaves}] Gravitational Wave Power} & \textcolor{teal}{$\text{\checkmark}$}  &  \textcolor{red}{$\text{\xmark}\phantom{0}$}  & \textcolor{red}{$\text{\xmark}$} & \textcolor{red}{$\text{\xmark}$} &  \textcolor{red}{$\text{\xmark}$} \\
    R & [\ref{ssec:einstein}] Relat. Time Dilat.  & \textcolor{teal}{$\text{\checkmark}$}  & \textcolor{red}{$\text{\xmark}\phantom{0}$} & \textcolor{red}{$\text{\xmark}$}& \textcolor{red}{$\text{\xmark}$}&  \textcolor{red}{$\text{\xmark}$} \\
    R & [\ref{ssec:kepler}] Kepler's 3 Law & \textcolor{teal}{$\text{\checkmark}$}  & \textcolor{orange}{$\text{\checkmark}^\ast$}  & \textcolor{red}{$\text{\xmark}$} & \textcolor{teal}{$\text{\checkmark}$} &  \textcolor{red}{$\text{\xmark}$}  \\
    S & \textcolor{blue}{[\ref{ssec:bell}] Bell inequalities$^\dagger$} & \textcolor{teal}{$\text{\checkmark}$}  & \textcolor{red}{$\text{\xmark}\phantom{0}$} & \textcolor{red}{$\text{\xmark}$} & \textcolor{red}{$\text{\xmark}$} & \textcolor{red}{$\text{\xmark}$}  \\
    
    \midrule
    S & \textcolor{blue}{[\ref{ssec:fsrd}] {I.15.10} FSRD} & \textcolor{teal}{$\text{\checkmark}$} & \textcolor{orange}{$\text{\checkmark}^\ast$} & \textcolor{red}{$\text{\xmark}$} & \textcolor{red}{$\text{\xmark}$} & \textcolor{red}{$\text{\xmark}$}  \\
    S & \textcolor{blue}{[\ref{ssec:fsrd}] {I.27.6} FSRD} & \textcolor{orange}{\phantom{0}$\text{\checkmark}^{\triangleright}$}
    & \textcolor{orange}{$\text{\checkmark}^\ast$} & \phantom{0}\textcolor{orange}{$\text{\checkmark}^\ast$} & \textcolor{teal}{$\text{\checkmark}$} & \textcolor{red}{$\text{\xmark}$}  \\
    S & \textcolor{blue}{[\ref{ssec:fsrd}] {I.34.8} FSRD} & \textcolor{orange}{\phantom{0}$\text{\checkmark}^{\triangleright}$}
    & \textcolor{teal}{$\text{\checkmark}$} \phantom{0} & \phantom{0}\textcolor{orange}{$\text{\checkmark}^\ast$} & \textcolor{teal}{$\text{\checkmark}$}&  \phantom{0}\textcolor{orange}{$\text{\checkmark}^\ast$}  \\
    S & \textcolor{blue}{[\ref{ssec:fsrd}] {I.43.16} FSRD} & \textcolor{teal}{$\text{\checkmark}$} & \textcolor{orange}{$\text{\checkmark}^\ast$}  & \phantom{0}\textcolor{orange}{$\text{\checkmark}^\ast$} & \textcolor{teal}{$\text{\checkmark}$}& \phantom{0}\textcolor{orange}{$\text{\checkmark}^\ast$}  \\
    S & \textcolor{blue}{[\ref{ssec:fsrd}] {II.10.9}  FSRD} & \textcolor{teal}{$\text{\checkmark}$} &\textcolor{orange}{$\text{\checkmark}^\ast$} & \phantom{0}\textcolor{orange}{$\text{\checkmark}^\ast$}& \phantom{0}\textcolor{orange}{$\text{\checkmark}^\ast$} & \phantom{0}\textcolor{orange}{$\text{\checkmark}^\ast$}  \\
    S & \textcolor{blue}{[\ref{ssec:fsrd}] {II.34.2} FSRD} & \textcolor{teal}{$\text{\checkmark}$} & \textcolor{orange}{$\text{\checkmark}^\ast$}  & \phantom{0}\textcolor{orange}{$\text{\checkmark}^\ast$} & \phantom{0}\textcolor{orange}{$\text{\checkmark}^\ast$} & \phantom{0}\textcolor{orange}{$\text{\checkmark}^\ast$}   \\
    
    \midrule
    S & \textcolor{blue}{[\ref{append:relcollision}] Inelastic Relativ. Collision }
    & \textcolor{teal}{$\text{\checkmark}$}  & \textcolor{orange}{$\text{\checkmark}^\ast$}  &  \textcolor{red}{$\text{\xmark}$}
    & \textcolor{red}{$\text{\xmark}$} & \textcolor{red}{$\text{\xmark}$}  \\
    S & \textcolor{blue}{[\ref{ssec:pion}] Decay of Pion$^\diamond$}
    & \textcolor{teal}{$\text{\checkmark}$}  & \textcolor{orange}{$\text{\checkmark}^\ast$}  & \textcolor{red}{$\text{\xmark}$}
    & \textcolor{red}{$\text{\xmark}$} &  \phantom{0}\textcolor{orange}{$\text{\checkmark}^\ast$} \\
    S & \textcolor{blue}{[\ref{ssec:lightscattering}] Radiation Damping}
    & \textcolor{teal}{$\text{\checkmark}$}  & \textcolor{red}{$\text{\xmark}\phantom{0}$}  &  \textcolor{red}{$\text{\xmark}$}
    & \phantom{0}\textcolor{orange}{$\text{\checkmark}^\ast$} & \phantom{0}\textcolor{orange}{$\text{\checkmark}^\ast$}  \\
    S & \textcolor{blue}{[\ref{ssec:escapevelocity}] Escape Velocity}
    & \textcolor{teal}{$\text{\checkmark}$}  & \textcolor{teal}{$\text{\checkmark}\phantom{0}$}  &  \textcolor{orange}{$\text{\checkmark}^\ast$}
    & \phantom{0}\textcolor{orange}{$\text{\checkmark}^\ast$} & \phantom{0}\textcolor{orange}{$\text{\checkmark}^\ast$}  \\
    S & \textcolor{blue}{[\ref{ssec:halleffect}] Hall Effect}
    & \textcolor{teal}{$\text{\checkmark}$}  & \textcolor{orange}{$\text{\checkmark}^\ast$}  & \textcolor{red}{$\text{\xmark}$}
    & \phantom{0}\textcolor{orange}{$\text{\checkmark}^\ast$} &  \phantom{0}\textcolor{orange}{$\text{\checkmark}^\ast$} \\
    S & \textcolor{blue}{[\ref{ssec:compton}] Compton Scattering}
    & \textcolor{teal}{$\text{\checkmark}$}  &  \textcolor{orange}{$\text{\checkmark}^\ast$} & \textcolor{orange}{$\text{\checkmark}^\ast$}
    & \phantom{0}\textcolor{orange}{$\text{\checkmark}^\ast$} &  \phantom{0}\textcolor{orange}{$\text{\checkmark}^\ast$} \\
    \bottomrule  
  \end{tabular}}
\label{tab:comparison_scientificlaws}
\caption*{\color{black}
  \textbf{Notation used in table:}\\
  \textcolor{teal}{$\checkmark$} denotes the successful recovery of a scientific law. \\
  \textcolor{red}{\xmark} ~denotes the failure to recover a scientific law. \\
  \textcolor{orange}{$\checkmark^\ast$} denotes recovery up to constants, but not exact recovery. \\
  \textcolor{orange}{$\checkmark^\triangleright$} denotes successful recovery when some variables that are potentially not observable are still declared to be observable.\\
  $^\dagger$ Inequalities are not supported by the methods from the literature that we benchmarked against.\\
  $^\diamond$ To generate data for this problem, we varied the mass values of both the pion and the muon. These variations were intended to mimic small differences in the observed values of the masses, rather than altering the actual values across a broad range of values. 
  }

\end{table}

\newpage

\section{\color{black}Discovery of Additional Problems From Background Theory}\label{append:additionalexamples}

We validate \verb|AI-Hilbert| on six problems from Feynman's lectures, considered earlier in AI-Descartes~\cite{cornelio2021ai}. We also examined six additional problems to demonstrate \verb|AI-Hilbert|'s capacity to rediscover scientific laws from background theory alone. Specifically, in Appendix \ref{append:relcollision}, we derive the inelastic relativistic collision law, in Appendix \ref{ssec:pion}, we derive the kinetic energy and momentum when a pion at rest decays into a muon and a neutrino, in Appendix \ref{ssec:lightscattering} we derive the light scattering law, in Appendix \ref{ssec:escapevelocity} we derive the minimal velocity that enables an object to overcome the gravitational pull of a planet, in Appendix \ref{ssec:halleffect} we derive the Hall potential of an electrical conductor, and in Appendix \ref{ssec:compton} we derive the Compton scattering formula.

\subsection{Problems from FSRD}\label{ssec:fsrd}

We consider the background theory provided in in AI-Descartes~\cite{cornelio2021ai} for the following six problems from Feynman’s Lectures: I.15.10, I.27.6, I.34.8, I.43.16, II.10.9 and II.34.2. 
As shown in the previous table (Table \ref{tab:comparison_scientificlaws}), AI-Hilbert obtained the correct symbolic expression for each problem. For the second and third problems, we declared some potentially nonmeasurable variables measurable and permitted AI-Hilbert to search for polynomials incorporating these variables. However, the final polynomial found by AI-Hilbert did not include these nonmeasurable variables. 

\subsection{Inelastic Relativistic Collision}\label{append:relcollision}
We consider the problem of computing the momentum and speed of a particle created by colliding a fast-moving particle with a stationary particle. This problem appears within the exercises in the Feynman Lectures on Physics series \citep{feynman1965feynman}.

We first introduce relevant notation: let $m$ denote the mass of the moving particle, $c$ be the speed of light, $v_m:=4/5 c$ denote the speed of the moving particle pre-collision, $m_c$ and $v_c$ denote the mass and speed, respectively,  of the composite particle after the collision, $p_m$ and $p_c$ denote the momentum of the moving particle before and after collision, respectively, and $E_m, E_r, E_c$ denote, respectively, the energy of the moving particle before collision, the resting particle before collision and the composite particle after collision. 

Then, given that a particle of mass $m$ moving at a speed of $v_m=4c/5$ collides in elastically with a similar particle at rest, we have the relations
\begin{align*}
    p_c=\frac{4 m_c v_c c}{\sqrt{3(c^2-v_c^2)}}, m_c=\frac{4m}{\sqrt{3}}
\end{align*}
We now derive these relations axiomatically using \verb|AI-Hilbert|, from the following axioms 
(via appropriate polynomial multipliers):
\begin{align}
    p_m^2 (c^2-v_m^2) - m^2 v_m^2 c^2 =0 \label{eqn:momentummovingparticle}\\
    E_m^2 - (m c^2)^2 - p_m^2 c^2 =0 \label{eqn:squareenergy}\\
    E_r - m c^2=0 \label{eqn:energyrest}\\
    E_c^2 - m_c^2  c^4 - p_c^2 c^2 =0 \label{eqn:energycollision}\\
    2 E_m E_r - E_c^2 + E_m^2 + E_r^2=0 \label{eqn:conservationofenergysq}\\
    p_c - p_m=0 \label{eqn:conservationofmomentum}\\
    v_m^2 - 16/25 c^2 =0 \label{eqn:definevelocity}\\
    p_c^2 (c^2-v_c^2) - m_c^2 v_c^2 c^2=0. \label{eqn:definemomentumcomposite}
\end{align}

Equation \eqref{eqn:momentummovingparticle} gives the momentum of the moving particle pre-collision (by rewriting the expression $p_m = \frac{m v} {\sqrt{1-v^2/c^2}}$ as a polynomial equation). Equation \eqref{eqn:squareenergy} models the energy of the moving particle pre-collision, Equation \eqref{eqn:energyrest} models the energy of the particle at rest pre-collision, Equation \eqref{eqn:energycollision} models the energy of the combined particle post-collision, Equation \eqref{eqn:conservationofenergysq} models the conservation of energy (squaring all terms to reduce the degree of the proof certificate; we could instead introduce $E_c-E_m-E_r=0$ and obtain essentially the same result), Equation \eqref{eqn:conservationofmomentum} models the conservation of momentum, Equation \eqref{eqn:definevelocity} defines the velocity of the moving particle pre-collision (again squaring both sides to reduce the degree of the proof certificate), and Equation \eqref{eqn:definemomentumcomposite} defines the momentum of the composite particle. 

By requiring that only the terms $m, m_c, c$ appear in our final formula, \verb|AI-Hilbert| provides the following polynomial equality as an output
\[
-1.33 m_c^4 (c^2)^3 + 5.33 m^2 m_c^2 (c^2)^3 +9.48 m^4 (c^2)^3 = 0
\]
Moreover, this equality can be manipulated to rediscover the two relations at the start of this section. In particular, it follows directly from the equality that $m_c^2 = 16/3 m^2$, and hence $m_c = \frac{4m}{\sqrt{3}}$

Moreover, substituting this expression for $m_c$ into Equation \eqref{eqn:definemomentumcomposite} and re-arranging yields:
 \[
 p_c = \frac{16 m  v_c c}{3\sqrt{c^2-v_c^2}}.
\]

\subsection{Decay of Pion into Muon and Neutrino}\label{ssec:pion}
Next, we consider the problem of deriving the kinetic energy and momentum generated when a pion at rest decays into a muon and a neutrino. 

We first introduce relevant notation: let $m_{\pi}, m_{\mu}$ and $m_{\nu}$ stand for, respectively, the mass of the pion, muon, and neutrino. Further, let $p_\mu$ be the momentum of the muon, $p_\nu$ be the momentum of the neutrino, and let $E_\pi, E_\mu, E_\nu$ be the respective total energy of the pion, the muon, and the neutrino. Given these quantities, the momentum of the neutrino satisfies the relation
\begin{equation}
p_{\nu}  = \frac{m_{\pi}^2 - m_{\mu}^2} {2 m_\pi} .\label{eqn:pionmuonneutrino}
\end{equation}

Accordingly, we now derive this relation by adding together the following background theory axioms with appropriate polynomial multipliers (normalizing in order that $c=1$ for simplicity):
\begin{eqnarray}
    p_{\nu} - p_{\mu}=0 \label{eqn:PION_CONSMOMENTUM}\\
    E_{\pi} - m_{\pi}=0 \label{eqn:defineenergyofpion}\\
    E_{\nu} - p_{\nu}=0 \label{eqn:momentumofmuon}\\
    E_{\pi} - E_{\mu} - E_{\nu}=0 \label{eqn:consenergy}\\
    E_{\mu}^2 - p_{\mu}^2 - m_{\mu}^2=0 \label{eqn:kinematicenergymuon}
\end{eqnarray}
where Equation \eqref{eqn:PION_CONSMOMENTUM} ensures that the decay obeys conservation of momentum, Equation \eqref{eqn:defineenergyofpion} defines the energy of the pion (recall that we set $c=1$ here), Equation \eqref{eqn:momentumofmuon} defines the energy of the neutrino (which equals its momentum as it is massless), Equation \eqref{eqn:consenergy} enforces conservation of energy, and Equation \eqref{eqn:kinematicenergymuon} defines the kinematic equation of the muon particle. 

We assume that the only measured quantities are $p_\nu, m_\pi, m_\mu$. Then \verb|AI-Hilbert| derives the equality
\[ 
-2m_{\pi}^2 + 2 m_{\mu}^2  + 4 p_{\nu} m_{\pi}=0
\]
which is equivalent to the original relation \eqref{eqn:pionmuonneutrino}.

\subsection{Radiation Damping and Light Scattering}\label{ssec:lightscattering}
We now consider the problem of computing the total amount of energy  radiated by a non-relativistic acceleration of a charge.

Denote by $S$ the amount of energy that passes per square meter per second through a surface normal to the 
radiation, let $q_c$ represent an electric charge accelerating at rate $a_p$, let $r$ and $\theta$ represent, respectively, the distance and angle at which the radiation is observed, and suppose that 
the charge is oscillating with displacement $x_0$ at frequency $w$. 

Given this notation, we have the natural law:
\begin{align}
P = \frac{4}{3} \pi q_c^2 x_0^2 w^4\label{eqn:radiationdamping}
\end{align}

For this problem, we use the following background theory:
\begin{eqnarray}
    S  r^2 - q_c^2  a_p^2  sin(\theta)^2 =0 
    \label{eqn:scaterring_rate}\\
    dA - 2 \pi r^2  sin(\theta)  d\theta =0
    \label{eqn:scaterring_area}\\
    P - \int_0^{\pi}S  dA = 0
    \label{eqn:scaterring_power2rate}\\
    \frac{4}{3} -  \int_0^{\pi} sin(\theta)^3  d\theta =0
     \label{eqn:scaterring_defint}\\
     a_p^2 - \frac{1}{2} w^4  x_0^2 = 0
     \label{eqn:scaterring_acceleration} 
    \end{eqnarray}
where Equation \eqref{eqn:scaterring_rate} models the rate of radiation of energy, Equation \eqref{eqn:scaterring_area} represents a differential of the area of a spherical segment, Equation \eqref{eqn:scaterring_power2rate} is the power to rate of radiation relation, Equation \eqref{eqn:scaterring_defint} is a definite integral from $0$ to $\pi$, such an integral can be provided as background knowledge, or can be solved by symbolic/numerical integrator, and lastly Equation \eqref{eqn:scaterring_acceleration} refers to the average of the acceleration of an oscillating charge over a cycle (squared).

\verb|AI-Hilbert| provides the relation
\[ 
4.0 P - 16.76 q_c^2 w^4 x_0^2 =0
\]
which can easily be manipulated to recover Equation \eqref{eqn:radiationdamping}.

\subsection{Escape Velocity}\label{ssec:escapevelocity}
We now consider the problem of deriving the escape velocity of a sphere-shaped planet, that is, the minimal velocity needed for any object attempting to exit a planet to overcome the planet's gravitational pull. Let $M$ denote the mass of the planet, $r$ denote its radius, $m$ denote the mass of the object, $v_e$ denote the escape velocity of the object, $K_i$ and $K_f$ denote the initial and final kinetic energy of the object, $U_i$ and $U_f$ denote the initial and final potential energy of the system, and $G$ denote the universal gravitational constant. We have the relation:

\begin{align}
    v_e =  \sqrt{\frac{2 G M}{r}}\label{eqn:escapevelocity}
\end{align}

We now recover this scientific law via \verb|AI-Hilbert| from the background theory consisting of the following polynomial equalities:
\begin{eqnarray}
    K_i - \frac{1}{2} m v_e^2 = 0
    \label{eqn:escape_initialk}\\
    K_f - 0 = 0 
     \label{eqn:escape_finalk}\\
U_i r + G M  m = 0 
      \label{eqn:escape_initialg}\\
U_f - 0 = 0       \label{eqn:escape_finalg}\\
K_i + U_i - (K_f + U_f) = 0, 
\label{eqn:escape_conservation}
\end{eqnarray}
where Equation \eqref{eqn:escape_initialk} models the initial kinetic energy of the object, Equation \eqref{eqn:escape_finalk} models the final kinetic energy of the object, Equations \eqref{eqn:escape_initialg}--\eqref{eqn:escape_finalg} reflect the initial and final potential energy of the body, and Equation \eqref{eqn:escape_conservation} models conservation of energy. 

Assuming that the only measured quantities are $G, M, m, r$ and $v_e$.
\verb|AI-Hilbert| uncovers the following polynomial relation that is derivable from the background theory:
\[
2 G M m - m v_e^2 r = 0.
\]
This can easily be rearranged to recover Equation \eqref{eqn:escapevelocity}.

\subsection{Hall Effect}\label{ssec:halleffect}

The Hall effect produces a potential difference (the Hall voltage, $U_H$) across an electrical conductor transverse to an electric current in the conductor and to an applied magnetic field perpendicular to the current.
Let a metal plate of length $L$, width $h$, and depth $d$ be subjected to a potential field across its length dimension. The plate is placed in a homogeneous magnetic field $B$ and perpendicular to the electric field $E$ (oriented with the depth dimension).
 The electron charges  $q_e$ moving at velocity $v$ across the length dimension, will be deflected by the magnetic field (due to a Lorenz force $F_m$). This will entail an excess of electrons on one side of the plate, and a deficiency on the other side, resulting in an electric field $E$ and respective electric force $F_e$ opposite to the magnetic force. Let the potential difference across the width dimension of the plate be $U_H$. Finally, let $I$ denote the current, which is defined by the number of electron charges $N$ crossing per time interval $dt$.
The Hall potential is given by the expression:
\begin{align}\label{eqn:hallpotential}
   U_H = \frac{h L I B}{ N q_e}. 
\end{align}

Equation \eqref{eqn:hallpotential} is recoverable from the following background theory via AI-Hilbert:
\begin{eqnarray}
    F_m - q_e  v  B = 0
\label{eqn:hall_lorenz_force} \\
    F_e - q_e  E = 0
\label{eqn:hall_electric_force} \\
    F_m - F_e = 0
\label{eqn:hall_newton3rd} \\
    E  h - U_H = 0
\label{eqn:hall_field2potential} \\
    v  dt - L = 0
\label{eqn:hall_velocity} \\
    I  dt - N  q_e = 0
    \label{eqn:hall_charge2current} 
\end{eqnarray}
where Equation \eqref{eqn:hall_lorenz_force} models the Lorentz magnetic force, Equation \eqref{eqn:hall_electric_force} represents the electric force ($E$ is the homogeneous electric field between the upper and lower metal plate), Equation \eqref{eqn:hall_newton3rd} models Newton's third law, Equation \eqref{eqn:hall_field2potential} relates the electric potential $U_H$ to field $E$  per plate of width $h$, Equation \eqref{eqn:hall_velocity} models the traversal velocity $v = \frac{L}{dt}$ across the plate length $L$, and Equation \eqref{eqn:hall_charge2current} models the amount of charge Q, is the current I times $dt$ (can be measured by an amp meter).

We assume the measured variables are $L, h, I, B, N, q_e$ and $U_H$.
\verb|AI-Hilbert| provides the relation
\[
q_e^2 U_H L N^2 - q_e B h L^2 I N=0,
\]
which can easily be rearranged to recover Equation \eqref{eqn:hallpotential}.

\subsection{Compton Scattering}\label{ssec:compton}

Compton scattering refers to the behavior of high frequency photons as they scatter after collision with a charged particle, usually an electron e in an atom. Specifically, when a photon knocks out a loosely bound electron from the outer valence shells of of an atom (or molecule), a new photon is emitted from the atom traveling at an angle $\theta$ to the incoming photon's path. Compton related the shift in photon wavelengths to the scattering angle: $\lambda_2 - \lambda_1 = \frac{h}{m_ec}(1-\cos \theta)$ where $\lambda_1, \lambda_2$ are the initial and final photon wavelengths, $m_e$ is the electron rest mass, $c$ is the speed of light, and $h$ is Planck's constant.

The following equations give a complete set of axioms needed to derive Compton's formula.
Let $E_r$ be the energy of the electron at rest (i.e., an electron bound to an atom), let $E_m$ be the energy of the moving electron (after it is knocked out from the atom).
Let $E_1, E_2$ be the initial and final photon energies, $f_1, f_2$ be the initial and final photon frequencies, and $p_1, p_2$ be the initial and final photon momentum values. The first constraint is just the expression for conservation of energy. The second and third constraints give the photon energy in terms of frequency. The subsequent two constraints give photon momentum in terms of frequency, and the two constraints after that relate wavelength and frequency. The next constraint is the mass-energy equivalence of the electron at rest, while the constraint after that gives the energy of the moving electron via the  relativistic energy–momentum relation. The last equation is a restatement of conservation of momentum (after squaring). 

\begin{eqnarray}
    &E_1+E_r-E_2-E_m &= 0 \\ 
    &E_1-hf_1 &= 0  \\
    &E_2-hf_2 &= 0 \\
    &p_1c - hf_1 &= 0 \\
    &p_2c - hf_2 &= 0 \\
    &\lambda_1f_1-c &= 0 \\
    &\lambda_2f_2-c &= 0 \\
    &E_r-m_ec^2 &= 0\\
    &E_m^2-pe_2^2c^2-m_e^2c^4 &= 0 \\
    &pe_2^2-p_1^2-p_2^2+2p_1p_2\cos \theta &= 0
\end{eqnarray}    

If we assume that the only measured variables are $h, c, m_e, \lambda_1, \lambda_2$ and $\cos \theta$,
AI Hilbert provides the relationship
\[ h^2c^2 - h^2c^2\cos^2 \theta - \lambda_2hm_ec^3 + \lambda_1hm_ec^3 - \lambda_2hm_ec^3\cos \theta + \lambda_1hm_ec^3\cos \theta = 0, \]
which becomes, after rearranging terms
\[ (\lambda_2 - \lambda_1)hm_ec^3(1+\cos \theta) = h^2c^2 (1-\cos^2\theta).\]
Dividing both sides by $hm_ec^3(1+\cos \theta)$, we get the desired relationship. An important assumption we made above is that measurements of $\cos \theta$ are available and not just of $\theta$. This is reasonable in many areas of physics as we often take the component of velocity, force etc. in a direction at an angle $\theta$ to the direction of the velocity, force etc. It is also reasonable to assume that $h, c, m_e$ are measured quantities, as these are known quantities.

\end{document}